\documentclass{article}

\usepackage[colorlinks=true,citecolor=brown,urlcolor=gray,linkcolor=BlueViolet]{hyperref}

\usepackage[preprint]{icml2026}

\usepackage{amsmath}
\usepackage{amssymb}
\usepackage{mathtools}
\usepackage{amsthm}

\usepackage{mathtools}

\usepackage{microtype}
\usepackage{graphicx}
\usepackage{subcaption}
\usepackage{booktabs} 

\usepackage{enumitem}
\usepackage{arydshln} 
\usepackage[utf8]{inputenc} 
\usepackage[T1]{fontenc}    

\usepackage{url}            
\usepackage{wrapfig,lipsum,booktabs}       
\usepackage{amsfonts}       
\usepackage{nicefrac}       
\usepackage[dvipsnames,table]{xcolor}  

\usepackage{colortbl}
\usepackage{pifont}
\usepackage{rotating}
\usepackage{makecell}

\usepackage{bbm}
\usepackage[most]{tcolorbox}
\usepackage{caption}
\usepackage{subcaption}
\usepackage{xspace}
\usepackage{multirow}

\usepackage[titletoc]{appendix}

\usepackage{pdflscape}

\usepackage{tabularray}

\usepackage{soul}

\usepackage[capitalize,noabbrev]{cleveref}

\usepackage[textsize=tiny]{todonotes}

\icmltitlerunning{Distributional Distillation for One-Pass Uncertainty Estimation}

\newcommand{\cmark}{\ding{51}}%
\newcommand{\xmark}{\ding{55}}

\newcommand{\ours}{\texttt{Dist$^2$ill}\xspace}
\newcommand{\llamaoneb}{\texttt{Llama3-1B-Instruct}\xspace}
\newcommand{\llamaeightb}{\texttt{Llama3-8B-Instruct}\xspace}
\newcommand{\qwenthreeb}{\texttt{Qwen2.5-3B-Instruct}\xspace}
\newcommand{\qwenmathoneb}{\texttt{Qwen2.5-Math-1.5B-Instruct}\xspace}
\newcommand{\qwenmathsevenb}{\texttt{Qwen2.5-Math-7B-Instruct}\xspace}
\newcommand{\qwenfourb}{\texttt{Qwen3-4B Non-thinking}\xspace}
\newcommand{\qweneightb}{\texttt{Qwen3-8B Non-thinking}\xspace}
\newcommand{\gemmaoneb}{\texttt{Gemma-3-1B-IT}\xspace}

\newcommand{\tfb}{TFB\xspace}
\newcommand{\blob}{BLoB\xspace}
\newcommand{\loraens}{ENS\xspace}
\newcommand{\loramle}{MLE\xspace}
\newcommand{\loramap}{MAP\xspace}
\newcommand{\loramcd}{MCD\xspace}
\newcommand{\loralap}{LAP\xspace}

\newlength\savewidth

\newcolumntype{C}{>{\centering\let\newline\\\arraybackslash\hspace{0pt}}m{2cm}}

\usepackage{amsmath,amsfonts,bm,eqnarray}
\usepackage{cancel}
\usepackage{amsthm}
\usepackage{tikz}
\usetikzlibrary{tikzmark}

\newtheorem{innercustomgeneric}{\customgenericname}
\providecommand{\customgenericname}{}
\newcommand{\newcustomtheorem}[2]{%
  \newenvironment{#1}[1]
  {%
   \renewcommand\customgenericname{#2}%
   \renewcommand\theinnercustomgeneric{##1}%
   \innercustomgeneric
  }
  {\endinnercustomgeneric}
}

\newcustomtheorem{customThm}{Theorem}
\newcustomtheorem{customLemma}{Lemma}
\newcustomtheorem{customCor}{Corollary}
\newcustomtheorem{customProposition}{Proposition}

\def\Tabref#1{Table~\ref{#1}}

\def\Figref#1{Fig.~\ref{#1}}

\def\appref#1{Appendix~\ref{#1}}
\def\Secref#1{Sec.~\ref{#1}}

\def\eqref#1{equation~\ref{#1}}
\def\Eqref#1{Eqn.~\ref{#1}}

\def\Algref#1{Algorithm~\ref{#1}}

\def\1{\bm{1}}

\def\rvx{{\mathbf{x}}}

\def\vtheta{{\bm{\theta}}}

\def\vh{{\bm{h}}}

\def\vp{{\bm{p}}}

\def\vs{{\bm{s}}}

\def\vx{{\bm{x}}}
\def\vy{{\bm{y}}}
\def\vz{{\bm{z}}}

\def\vphi{{\boldsymbol{\phi}}}
\def\vpi{{\boldsymbol{\pi}}}

\def\mP{{\bm{P}}}

\def\mS{{\bm{S}}}

\def\mX{{\bm{X}}}
\def\mY{{\bm{Y}}}

\DeclareMathAlphabet{\mathsfit}{\encodingdefault}{\sfdefault}{m}{sl}
\SetMathAlphabet{\mathsfit}{bold}{\encodingdefault}{\sfdefault}{bx}{n}

\def\gD{{\mathcal{D}}}

\def\gL{{\mathcal{L}}}

\newcommand{\E}{\mathbb{E}}

\DeclareMathOperator*{\argmax}{arg\,max}

\renewcommand{\tilde}{\widetilde}
\renewcommand{\hat}{\widehat}
\renewcommand{\frac}{\tfrac}

\definecolor{lightergray}{HTML}{e5e5e5}
\definecolor{lightgray}{HTML}{cccccc}
\sethlcolor{lightergray}
\sethlcolor{lightgray}

\begin{document}

\twocolumn[
  \icmltitle{\ours: Distributional Distillation for One-Pass Uncertainty \\ Estimation in Large Language Models}



  \icmlsetsymbol{equal}{*}
  \icmlsetsymbol{lead}{$\dagger$}

  \begin{icmlauthorlist}
    \icmlauthor{Yicong Zhao}{equal,ru}
    \icmlauthor{King Yeung Tsang}{equal,ru}
    \icmlauthor{Harshil Vejendla}{ru}
    \icmlauthor{Haizhou Shi}{lead,ru}
    \icmlauthor{Zhuohang Li}{lead,vu}
    \icmlauthor{Zhigang Hua}{meta}
    \icmlauthor{Qi Xu}{meta}
    \icmlauthor{Tunyu Zhang}{ru}
    \icmlauthor{Yi Wang}{ru}
    \icmlauthor{Ligong Han}{rh}
    \icmlauthor{Bradley A. Malin}{vu}
    \icmlauthor{Hao Wang}{ru}
  \end{icmlauthorlist}

  \icmlaffiliation{ru}{Rutgers University}
  \icmlaffiliation{vu}{Vanderbilt University}
  \icmlaffiliation{meta}{Meta}
  \icmlaffiliation{rh}{Red Hat AI Innovation}

  \icmlcorrespondingauthor{Haizhou Shi}{haizhou.shi@rutgers.edu}
  \icmlcorrespondingauthor{Zhuohang Li}{zhuohang.li@vanderbilt.edu}
  \icmlcorrespondingauthor{Hao Wang}{hw488@cs.rutgers.edu}

  \icmlkeywords{Machine Learning, ICML}

  \vskip 0.3in
]



\printAffiliationsAndNotice{
\icmlEqualContribution
$^\dagger$Project Co-Lead
}  


\begin{abstract}
Large Language Models (LLMs) often exhibit misalignment between the quality of their generated responses and the confidence estimates they assign to them. Bayesian treatments, such as marginalizing over a reliable weight posterior or over the space of reasoning traces, provide an effective remedy, but incur substantial computational overhead due to repeated sampling at test time.
To enable accurate uncertainty estimation in a single forward pass, we propose a novel \textbf{dist}ributional \textbf{distill}ation framework (\ours) that trains an LLM to produce multiple diverse reasoning paths within one inference pass, while using a lightweight parametric module to approximate empirical confidence scores derived from the sampling distribution. Extensive experiments demonstrate that \ours preserves reasoning diversity and achieves state-of-the-art uncertainty estimation, substantially improving Expected Calibration Error (ECE) and Negative Log-Likelihood (NLL), while remaining computationally efficient.
\end{abstract}

\section{Introduction}
\label{sec:intro}
Large Language Models (LLMs) have demonstrated remarkable reasoning capabilities across a wide range of complex domains~\cite{touvron2023llama,guo2025deepseek,yang2025qwen3,zeng2025glm}. 
Despite these advances, their deployment in high-stakes applications remains fundamentally limited by the lack of reliable confidence estimation~\cite{xiong2023can,huang2025survey,liu2025uncertainty,he2025survey,heo2024llms,kapoor2024large,zhou2023navigating,kadavath2022language}.
A commonly adopted approach to this issue is verbalized confidence estimation, where the model is directly prompted to state how confident it is in its answer~\cite{xiong2023can,yang2024verbalized,lin2022teaching}. However, extensive empirical evidence shows that such self-reported confidence is often severely miscalibrated, particularly in challenging reasoning settings~\cite{tian2023just,zhang2025tokur}. Moreover, verbalized confidence does not constitute a well-defined probability distribution over plausible outputs, which fundamentally limits its applicability in risk-sensitive decision-making~\cite{wang2026llm,tao2025revisiting,lee2024selective}.
These limitations have motivated sampling-based interfaces for LLMs. Rather than relying on a single response accompanied by a self-reported confidence score, this paradigm approximates the marginal distribution over final answers, $\pi_\vtheta(\vy\mid\vx){=}\E_{\vz\sim\pi_{\vtheta}(\vz\mid \vx)}[\pi_\vtheta(\vy\mid\vz,\vx)]$, by sampling multiple reasoning trajectories. The most confident prediction is then selected as $\vy^*{=}\argmax_{\vy} \pi_{\vtheta}(\vy \mid \vx)$, which has been shown to yield substantially more reliable outcomes in practice~\cite{wang2022self,stiennon2020learning,kang2025scalable,wang2025ranked,taubenfeld2025confidence}.

Inspired by the partial success of prior sampling-based approaches, we first provide strong empirical evidence for a recurring phenomenon that we term \emph{``Internal Alignment of Uncertainty~(IAU)''} in verifiable reasoning tasks~\cite{cobbe2021training,chen2021evaluating,rein2024gpqa}. Specifically, we show that the marginal predictive distribution over final answers, approximated by sampling multiple reasoning traces $\pi_\vtheta(\vy\mid\vx)$ exhibits reliable calibration when the number of samples is sufficiently large. Notably, this behavior consistently holds across different model families and architectures, despite their varying benchmark performance levels~(\Secref{sec:exp_teacher}).

However, directly exploiting this property as in test-time scaling~\cite{muennighoff2025s1,wang2025sampling,zhao2025sample,agarwal2025art} or Bayesian LLMs~\cite{shi2024training,yang2023bayesian,marszalek2025minimal,tonolini2024bayesian,zhang2025amortized} incurs prohibitive computational and memory costs, severely limiting practical deployment. To address this challenge, we propose a unified \textbf{Dist}ributional \textbf{Distill}ation framework, \textbf{\ours,} that compresses trace- or weight-level posterior marginalization into a single-pass inference process.
The core idea is to amortize the expensive sampling of reasoning traces or model weights by shifting this computational burden from inference to training, enabling efficient deployment under strict one-call budgets.
Concretely, \ours operates as an extended supervised fine-tuning~(SFT) stage for LLM post-training. We first sample multiple reasoning traces or model weights from a reference policy $\pi_{\vtheta}(\cdot \mid \vx)$ or an approximate weight posterior $q(\cdot\mid\vtheta)$ and aggregate the resulting predictions into an empirical categorical answer distribution. This distribution is then used as supervision, training the model to output multiple candidate solutions together with a calibrated probability allocation in a single forward pass.
In contrast to verbalized uncertainty estimation, where probabilities are generated solely through self-reporting without a principled grounding, our supervision is anchored in the model’s own empirical answer distribution obtained via explicit trace or weight marginalization.

We evaluate the effectiveness of \ours on a diverse set of commonsense and mathematical reasoning tasks, covering two downstream application paradigms (discriminative and generative) across multiple model architectures (\texttt{Qwen} and \texttt{Llama} models). Comprehensive experiments show that our distributional self-distillation framework consistently achieves stronger generalization and more accurate uncertainty estimation at substantially lower computational cost than widely used training-free baselines.
In summary, our main contributions are as follows:
\begin{itemize}[nosep,leftmargin=12pt]
    \item We empirically identify and systematically validate the phenomenon of \emph{``Internal Alignment of Uncertainty~(IAU)''} in LLMs, demonstrating that the approximated predictive distribution over final answers is well calibrated across diverse model families and scales.
    \item We introduce a novel distributional self-distillation framework that leverages trace- or weight-induced empirical answer distributions as supervision, enabling the LLM to produce multiple candidate solutions with an explicit probability allocation over final answers in a single inference pass.
    \item Through extensive experiments, we show that \ours consistently improves uncertainty estimation and downstream generalization while significantly reducing inference-time cost, highlighting its potential as a practical post-training paradigm for modern LLMs.
\end{itemize}

\section{Related Work}
\label{sec:related}
\textbf{Training-Free Uncertainty Estimation of LLMs.}\quad
From a probabilistic perspective, reliable uncertainty estimation hinges on obtaining a calibrated predictive distribution, typically achieved by marginalizing over latent sources of randomness.
Current training-free approaches seek to capture this uncertainty through two primary avenues.
In the parameter space, Training-Free Bayesianization (TFB)~\cite{shi2024training} transforms a trained LoRA into a Bayesian adapter without requiring extra training. In the output space, research has focused on Verbalized Uncertainty, where models are explicitly prompted to state their confidence~\cite{tian2023just,xiong2023can,wang2022self}, and Semantic Entropy, which aggregates uncertainty over clusters of semantically equivalent answers~\cite{kuhn2023semantic,lin2023generating,cole2023selectively}. 
However, these methods typically involve sampling multiple parameter sets or responses, introducing significant overhead at inference time.


More closely related to our work,
Verbalized Sampling~(VS)~\cite{zhang2025verbalized} introduces a training-free prompting strategy that instructs the model to generate multiple candidates with associated verbalized probabilities, aiming to mitigate mode collapse while yielding an explicit distribution over potential answers.
Given its ability to generate multiple solutions along with estimated probabilities in a single pass, VS serves as a compelling baseline for our study.
Yet, the reliability of such \emph{verbalized} probabilities in forming a stable and calibrated predictive distribution $\pi_\vtheta(\vy\mid \vx)$ for reasoning tasks has not been fully established, which prompts our empirical investigation.

\textbf{Training-Based Uncertainty Estimation of LLMs.}\quad
Training-based approaches offer a robust pathway for enhancing uncertainty estimation in LLMs.
A significant body of prior work investigates Bayesian model averaging and distillation-based approaches to approximate calibrated predictive distributions~\cite{hoeting1999bayesian, hinton2015distilling, korattikara2015bayesian, nemani2025efficient}.
More recent frameworks have leveraged supervised and reinforcement learning to align model probabilities with response correctness. These strategies encompass a diverse range of methods, from generative verifiers~\cite{li2025judging} and knowledge-based rejection~\cite{xu2024rejection} to calibrated reward modeling and policy optimization variants (e.g., PPO-M)~\cite{stangel2025rewarding,leng2024taming}.
 
In contrast to existing training-based approaches, which primarily focus on refining a \emph{single} generated response, our goal is to capture the full landscape of uncertainty by producing multiple candidate solutions and their associated probability distribution in a single inference pass.

\begin{figure*}[t]
    \centering
    \includegraphics[width=1\linewidth]{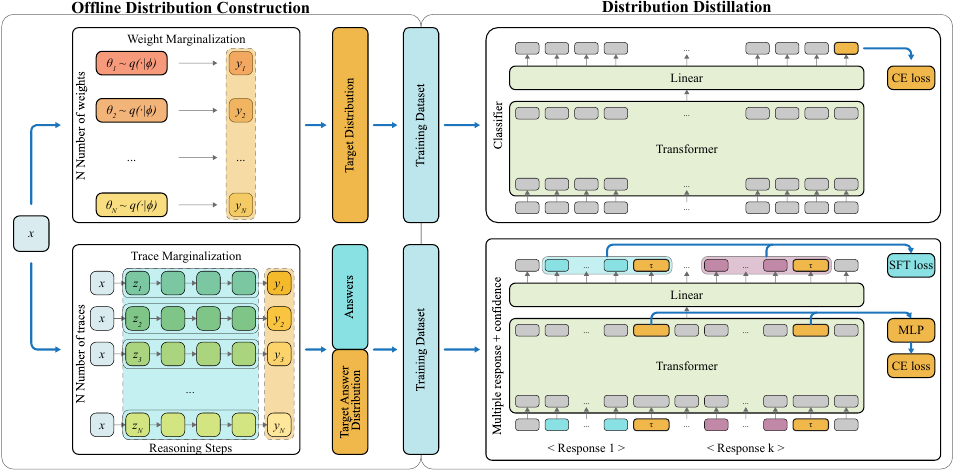}
    \caption{
    \textbf{Overview of our Two-Stage Distributional Distillation (\ours).}
    \textbf{(i) Offline distribution construction:} the predictive distributions are either constructed by \textit{weight marginalization} (for classification) or \textit{reasoning trace marginalization} (for generation) on the (unlabeled) training data; 
    \textbf{(ii) Distributional distillation:} using these predictive distributions as targets for distributional distillation. 
    }
    \label{fig:main}
    \vspace{-1em}
\end{figure*}

\section{Methodology}
\label{sec:methodology}

This section presents our proposed distributional distillation framework, \ours, for improved uncertainty estimation in LLMs.
\Secref{sec:method-preliminary} introduces the necessary preliminaries, discussing how marginalizing over model weights and reasoning traces enhances both predictive performance and uncertainty estimation.
\Secref{sec:method-main} provides an overview of \ours as a unified distributional distillation framework applicable to both sequence classification and sequence generation, with detailed formulations given in \Secref{sec:method-seq-classification} and \Secref{sec:method-seq-generation}, respectively.



\subsection{Preliminaries}
\label{sec:method-preliminary}

\textbf{Notation.}\quad
Throughout this paper, scalars are denoted by lowercase letters (e.g., $x$), vectors or token/word sequences by bold lowercase letters (e.g., $\vx$), random vectors by bold lowercase letters with a random-variable notation (e.g., $\rvx$), and matrices or collections of tokens, words, or phrases by bold uppercase letters (e.g., $\mX$). We use $[m]=\{1,2,\ldots,m\}$ to denote the set of consecutive integers from $1$ to $m$.
We use $\oplus$ to denote the string concatenation.
We follow the convention where $\pi_\vtheta(\cdot)$ denotes the conditional probability of the language model under training, while $q_{\vphi}(\cdot)$ represents the variational distribution.


\textbf{Bayesian LLMs (Weight Marginalization).}\quad
As a practical instantiation of Bayesian Neural Networks (BNNs)~\cite{tierney1986accurate,wang2020survey,blundell2015BBB,wang2016natural,kingma2015variational,mackay1992practical,bishop2006pattern,welling2011bayesian,ahn2012bayesian,izmailov2021bayesian}, Bayesian LLMs aim to approximate the true posterior over network parameters, $p(\vtheta \mid \gD)$, using a tractable variational distribution $q(\vtheta \mid \vphi)$, which is typically restricted to low-rank mean-field Gaussian families~\cite{yang2023bayesian,wang2024blob,shi2024training}.

By marginalizing over the approximate weight posterior, Bayesian LLMs yield substantially improved predictive distributions via Bayesian Model Averaging (BMA):
\begingroup\makeatletter\def\f@size{9.5}\check@mathfonts\def\maketag@@@#1{\hbox{\m@th\normalsize\normalfont#1}}
\begin{align}
\label{eq:bma}
    \nonumber p(\vy \mid \vx, \gD)
    &= \E_{\vtheta \sim p(\cdot \mid \gD)}\!\left[\pi_{\vtheta}(\vy \mid \vx)\right]
    \approx \E_{\vtheta \sim q(\cdot \mid \vphi)}\!\left[\pi_{\vtheta}(\vy \mid \vx)\right] \\
    &\approx \frac{1}{N} \sum\nolimits_{n=1}^{N} \pi_{\vtheta_n}(\vy \mid \vx),
    \quad \vtheta_n \sim q(\vtheta \mid \vphi).
\end{align}
\endgroup

Despite its proven effectiveness, exhibiting a positive correlation between test-time compute (i.e., the number of samples) and uncertainty estimation quality~\cite{wang2024blob,shi2024training}, deploying Bayesian LLMs in practice remains challenging. The multiple sampling passes required by \Eqref{eq:bma} impose substantial inference overhead, limiting their applicability primarily to narrow downstream classification tasks~\cite{yang2023bayesian,wang2024blob,shi2024training} or to generation settings with only an extremely small number of samples~\cite{zhang2025tokur}.
To address this limitation by avoiding multiple samples, our \ours distills a Bayesian teacher LLM into a single deterministic student model (\Secref{sec:method-seq-classification}).

\textbf{Self-Consistency (Reasoning Trace Marginalization).}\quad
For tasks with short-form, verifiable answers, LLM reasoning can be naturally formulated as marginalization over intermediate natural-language reasoning traces $\vz$ (i.e., sequences of tokens or words) to produce a distribution of final answers $\vy$. This marginalization can be approximated via Monte Carlo sampling~\cite{chen2024language,wang2022self}:
\begin{align}
    \label{eq:reasoning-as-integration}
    \pi_\vtheta(\vy \mid \vx)
    &= \E_{\vz \sim \pi_\vtheta(\cdot \mid \vx)} \!\left[\pi_\vtheta(\vy \mid \vz, \vx)\right] \\
    \nonumber &\approx \frac{1}{N} \sum\nolimits_{n=1}^{N} \pi_\vtheta(\vy \mid \vz_n, \vx),
    \quad \vz_n \sim \pi_\vtheta(\vz \mid \vx).
\end{align}
However, most existing approaches overlook this distributional perspective and instead adopt simplified variants. For example, setting $N{=}1$ reduces the formulation to standard Chain-of-Thought~(CoT) reasoning~\cite{wei2022chain}, while Majority Voting (Maj@N) partially approximates the marginal distribution but retains only the most frequent (confident) answer to improve accuracy.
In contrast, our work explicitly evaluates and exploits the approximated marginal distribution $\pi_\vtheta(\vy\mid \vx)$ to achieve improved alignment between answer quality and confidence~(\Secref{sec:method-seq-generation} and \ref{sec:exp_teacher}).


\subsection{\ours: Distributional Distillation of LLMs}
\label{sec:method-main}
\textbf{Assumptions.}\quad 
We assume we have access to a reliable predictive distribution $\pi(\vy\mid \vx)$, regardless of the task type or its specific form of modeling: it can be a sequence classification model designed for multi-choice question-answering (\Secref{sec:method-seq-classification}) or 
a causal generative model for solving complex reasoning tasks (\Secref{sec:method-seq-generation}). We further assume the quality of this LLM's reference signals is significantly better than the current student model $\pi_\vtheta$ we have. 

\textbf{Overview.}\quad
\Figref{fig:main} presents the overview of our proposed \ours framework and the complete workflow is summarized in \Algref{alg:ours}.

\begin{itemize}[nosep,leftmargin=12pt]
    \item \emph{For classification tasks:} we distill the ensemble-averaged posterior from Bayesian weight marginalization into a single student model, using a scheduled loss to stabilize soft-target approximation. 
    \item \emph{For generative tasks:} we compress trace-based marginalization into a single inference pass, training the student to predict multiple candidates and their aggregated probabilities within a structured output template. 
\end{itemize}

\subsection{\ours for Sequence Classification}
\label{sec:method-seq-classification}

\textbf{Training Objective.}\quad
Point estimates of LLM parameters after adaptation, particularly for sequence classification, often lead to severe misalignment between predictive quality and confidence estimates (e.g., low accuracy but high confidence)~\cite{yang2023bayesian,wang2024blob,shi2024training}. To mitigate this issue, we distill knowledge from a reliable Bayesian LLM teacher.

Let $q(\cdot \mid \vphi)$ denote the teacher’s approximate posterior over model parameters, where $\vphi$ parameterizes the weight distribution. 
Let $\vtheta$ denote the parameters of the deterministic student model. The training objective of \ours for classification~($\gL_{\text{cls}}$) is defined as
\begin{align}
\label{eq:kl-div}
    \gL_{\text{cls}}(\vtheta)
    &= \E_{\vx}\!\left[
        \gD_{\text{KL}}\!\left(
            \pi_{\vphi}(\vy \mid \vx)
            \,\mid\mid\,
            \pi_{\vtheta}(\vy \mid \vx)
        \right)
    \right],
\end{align}
where $\pi_{\vphi}(\vy\mid\vx) \triangleq \E_{\hat{\vtheta} \sim q(\cdot\mid\vphi)}[\pi_{\hat{\vtheta}}(\vy\mid\vx)]$ denotes the teacher’s marginalized predictive distribution, which is estimated via Monte Carlo sampling during training. 
We analyze the impact of the teacher’s sample size and alternative distillation losses in ~\appref{subsubsec:ablation_distillation_loss}.


\textbf{Training Details.}\quad
Directly training the student model using soft targets from the Bayesian teacher can be unstable. To improve training stability, we adopt a loss scheduling strategy that initially emphasizes learning the modal prediction (i.e., the highest-probability output) and gradually transitions to full predictive distribution matching~\appref{app:simple_loss_s}. In addition, for smaller datasets, we apply paraphrase-based data augmentation, as we find that limited data substantially affects the performance of \ours, which is a common setting in downstream classification tasks~\appref{app:data_aug}.

\subsection{\ours for Sequence Generation}
\label{sec:method-seq-generation}

\textbf{Sampling Distributional Responses.}\quad
The objective of our approach is to train a model that directly outputs a distribution over final answers, thereby avoiding costly repeated sampling at test time.

Concretely, for each input query $\vx$, we sample $N$ reasoning traces along with their induced final answers,
$(\vz_n, \vy_n) \overset{\text{i.i.d.}}{\sim} \pi_{\vtheta}(\cdot \mid \vx), n \in [N]$.
In practice, semantically equivalent but differently phrased final answers may arise; thus, we apply answer canonicalization to group identical final answers.
For notational simplicity, we assume throughout that all final answers are automatically canonicalized.
The trace-induced empirical distribution over final answers is then defined as
\begin{equation}
\hat{\pi}_{\vtheta}(\vy \mid \vx)
\triangleq \frac{1}{N} \sum\nolimits_{n=1}^{N} \mathbbm{1}_{[\vy_n = \vy]} .
\label{eq:trace-induced}
\end{equation}

However, even for sequence generation tasks with verifiable final answers, the space of valid answers can be extremely large and cannot be exhaustively enumerated. This poses a challenge for effective distributional modeling, as requiring the model to explicitly represent a vast number of low-probability answers is impractical. To address this issue, we further approximate the empirical distribution using a top-$K$ truncation.

Without loss of generality, we assume that the first $K$ sampled responses
$\{(\vy_k, \pi_k)\}_{k \in [K]}$
correspond to the $K$ most frequent final answers, where
$\pi_k \triangleq \hat{\pi}_{\vtheta}(\vy_k \mid \vx)$
denotes the estimated confidence of answer $\vy_k$, and the responses are ordered in descending confidence.
To capture the remaining long-tail uncertainty and ensure a valid probability simplex, we introduce an \emph{aggregated probability} of remaining answers
$(\vy_{K+1}, \pi_{K+1}) \triangleq (\text{``others''}, 1 - \sum_{k=1}^{K} \pi_k)$.

Since it is well established that LLMs exhibit improved robustness and reasoning ability when allowed to generate intermediate reasoning steps (i.e., ``think step by step''~\cite{kojima2022large}), we additionally resample reasoning traces $\vz$ during training according to the following empirical posterior:
\begingroup\makeatletter\def\f@size{9}\check@mathfonts\def\maketag@@@#1{\hbox{\m@th\normalsize\normalfont#1}}
\begin{equation}
\hat{\pi}_{\vtheta}(\vz \mid \vy, \vx) =
\begin{cases}
\displaystyle
\frac{1}{\sum_{n=1}^{N} \mathbbm{1}_{[\vy_n = \vy]}} ,
& \text{if } \vz \in \{\vz_n \mid \vy_n = \vy\}_{n \in [N]}, \\[6pt]
0, & \text{otherwise}.
\end{cases}
\end{equation}
\endgroup
Finally, the distributional responses used for our \ours are formalized as the set of triplets $\mS = \{(\vz_k', \vy_k, \pi_k)\}_{k\in [K+1]}$, where $\vz_k' \sim \hat{\pi}_{\vtheta}(\vz \mid \vy_k, \vx)$.

\textbf{Modeling and Training Objective.}\quad
Unlike sequence classification as discussed in \Secref{sec:method-seq-classification}, where predictions are made over a \emph{fixed and predefined} label space, sequence generation requires modeling an open-ended set of answer supports. Accordingly, our \ours explicitly approximates this open answer space and estimates confidence based on the generated responses themselves.

Inspired by recent architectures for process reward modeling~\cite{zhang2025lessons}, we decouple token generation from confidence estimation. Concretely, for an input query $\vx$ and its associated sampled triplet set
$\mS = \{(\vz_k', \vy_k, \pi_k)\}_{k \in [K+1]}$,
the response used for Supervised Fine-Tuning~(SFT) is constructed as
\begin{align}
    \vs(\vx, \mS)
    = \bigoplus\nolimits_{k=1}^{K+1} \bigl(\vz_k' \oplus \vy_k \oplus \tau\bigr),
\end{align}
where $\tau$ is a special delimiter token that separates different responses and serves as an anchor point for subsequent confidence prediction.\footnote{
Another formulation $\vs(\vx, \mS)
    = \bigoplus\nolimits_{k=1}^{K+1} \bigl(\vz_k' \oplus \vy_k \oplus \pi_k\bigr),$ incorporates the confidence estimation directly in the space of tokens, which reduces our confidence estimation module to verbalized uncertainty.
}
Hence the SFT loss for sequence generation in \ours is defined as
\begin{equation}
\label{eq:Lsft}
    \mathcal{L}_{\text{sft}}(\vtheta)
    = \E_{\vx, \mS}\!\left[
        - \log \pi_{\vtheta}\!\left(\vs(\vx, \mS) \mid \vx\right)
    \right].
\end{equation}
For confidence estimation, we define the target distribution
$\vpi(\mS) = [\pi_1, \ldots, \pi_{K+1}]^\top$
over the corresponding answer supports
$[\vy_1, \ldots, \vy_{K+1}]^\top$.
The predictive confidence distribution is produced by an auxiliary MLP head applied to the final-layer hidden state at the delimiter token $\tau$ in $\vs(\vx, \mS)$, and is denoted as $\pi_{\vtheta}(\vy \mid \vx, \mS)$.
The resulting cross-entropy-based calibration loss is
\begin{equation}
\label{eq:Lce}
    \mathcal{L}_{\text{ce}}(\vtheta)
    = \E_{\vx, \mS}\!\left[
        - \sum\nolimits_{k=1}^{K+1} \pi_k
        \log \pi_{\vtheta}(\vy_k \mid \vx, \mS)
    \right].
\end{equation}
The overall training objective for sequence generation tasks is a weighted combination of \Eqref{eq:Lsft} and \ref{eq:Lce},
\begin{align}
    \gL_{\text{gen}}(\vtheta)
    = \mathcal{L}_{\text{sft}}(\vtheta)
    + \lambda_t \mathcal{L}_{\text{ce}}(\vtheta),
\end{align}
where $\lambda_t$ follows a curriculum-based schedule (refer to \appref{app:imple_details} for details).

\section{Experiments}
\label{sec:experiments}
We evaluate \ours through extensive experiments.

\subsection{Settings}
\label{sec:settings}

\textbf{Models.}\quad 
We implement our method across different architectures and scales corresponding to the benchmarks: 
\emph{For sequence classification, }we follow the previous work~\cite{wang2024blob,shi2024training} using the latest open-source \texttt{Meta-Llama-3.1-8B}~\cite{dubey2024llama}.
\emph{For sequence generation,} we train \texttt{Qwen3-4B/8B Non-thinking} with their own trace-induced distributional supervision. 

\begin{figure*}[t!]
    \centering
    \includegraphics[width=\textwidth]{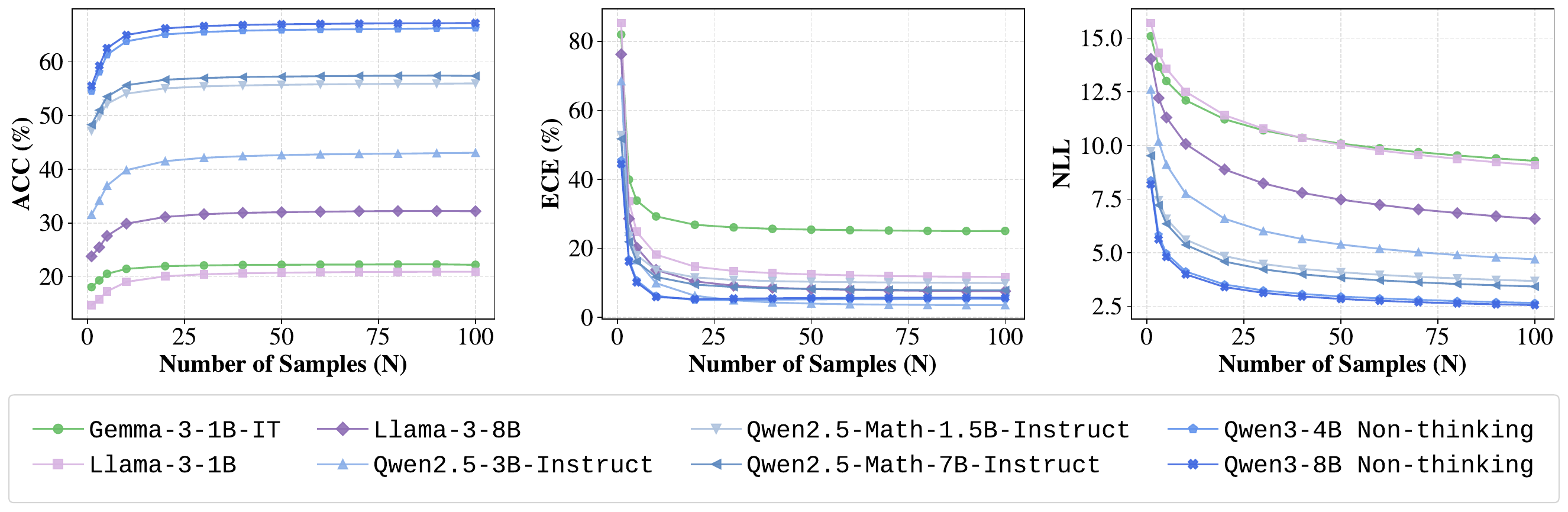}
    \caption{
        \textbf{Demonstration of Internal Alignment of Uncertainty~(IAU).} 
        We analyze the predictive performance and uncertainty quality of trace-induced empirical answer distributions $\pi_\vtheta(\vy\mid\vx)\approx \frac{1}{N}\sum\nolimits_{n=1}^N\pi_\vtheta(\vy|\vz_n,\vx),\vz_n\sim \pi_\vtheta(\cdot\mid \vx)$, as a function of the number of reasoning traces $N$. Across diverse model families, predictive accuracy quickly saturates, whereas uncertainty metrics (ECE and NLL) continue to improve, demonstrating that LLMs universally possess the property of \emph{Internal Alignment of Uncertainty}.
    }
    \label{fig:teacher-samples}
    \vspace{-1em}
\end{figure*}

\textbf{Benchmarks.}\quad 
Two types of benchmarks are used: 
\begin{itemize}[nosep,leftmargin=12pt]
    \item \emph{Sequence Classification:} we follow the existing literature of Bayesian LLMs and benchmark on six commonsense reasoning tasks, including Winogrande-Small~(\textbf{WG-S}) and Winogrande-Medium~(\textbf{WG-M})~\cite{wg}, ARC-Challenge~(\textbf{ARC-C}) and ARC-Easy~(\textbf{ARC-E})~\cite{arc}, Open Book Question Answering~(\textbf{OBQA})~\cite{obqa}, and BoolQ~\cite{boolq}. We further use the model trained on OBQA~\cite{obqa} to evaluate their generalization ability on out-of-distribution datasets: college-level chemistry~(\textbf{Chem}) and physics~(\textbf{Phy}) subsets of MMLU~\cite{mmlu}.
    \item \emph{Sequence Generation:} we mainly evaluate the methods on mathematical reasoning datasets as they require complex reasoning capability, including moderate \textbf{MATH-500}~\cite{hendrycks2021measuring} and challenging \textbf{AIME 24\&25}~\cite{aime}.
\end{itemize}

\textbf{Evaluation Metrics.}\quad 
To evaluate the uncertainty estimation quality of different methods, we measure Expected Calibration Error~(\textbf{ECE}~\cite{naeini2015obtaining}) and Negative Log-Likelihood~(\textbf{NLL}). We also report Accuracy~(\textbf{ACC}) to ensure models maintain strong performance. 
\emph{For sequence generation tasks,} we additionally report the metrics of answer diversity~(\textbf{Div}) and class-wise ECE~(\textbf{ECE}$_{\text{cw}}$)~\cite{kull2019beyond}, which measures calibration aggregated across all candidate slots.
Detailed metric definitions are provided in \appref{app:exp_settings_metrics}.

\textbf{Baselines.}\quad We compare \ours against strong baselines under the same evaluation protocol (refer to \appref{app:baseline_details} for details).
\begin{itemize}[nosep,leftmargin=12pt]
    \item \emph{For Sequence Classification:} we mainly include state-of-the-art Bayesian LLMs as baselines, including 
ensemble-based method: Deep Ensemble (\textbf{\loraens})~\cite{lakshminarayanan2017simple,balabanov2024uncertainty,wang2023lora}, 
variational inference methods: Monte-Carlo Dropout (\textbf{\loramcd})~\cite{gal2016dropout}, 
Bayesian LoRA by Backprop~(\textbf{\blob})~\cite{wang2024blob}, and post-training method: Laplace-LoRA~(\textbf{\loralap})~\cite{yang2023bayesian} and Training-Free Bayesianization~(\textbf{\tfb})~\cite{shi2024training}.
We also include three non-Bayesian methods~(point estimation of the model weights), including two standard PEFT baselines: Maximum Likelihood Estimation~(\textbf{MLE})~\cite{hu2022lora} and Maximum A Posteriori~(\textbf{MAP}), and the weight mean of \blob (\textbf{\blob-M}).
    \item \emph{For Sequence Generation:} as directly predicting the output distribution within a single-pass of LLMs emerges as a relatively novel direction, with not much prior work to discuss, we include Chain-of-Thought~(\textbf{CoT})~\cite{wei2022chain}, Verbalized Sampling~(\textbf{VS})~\cite{zhang2025verbalized}, Multi-turn Sampling~(\textbf{MT}), and Parallel Sampling~(PS)~\cite{brown2024large} as the baseline methods.
\end{itemize}

\textbf{Implementation Details.}\quad
\ours is implemented as a two-stage pipeline consisting of
\emph{(i) offline distribution construction} and
\emph{(ii) distributional distillation}.
Due to space constraints, we defer implementation details to \appref{app:imple_details}.

\subsection{Empirical Study: Internal Alignment of Uncertainty in LLMs (IAU)}
\label{sec:exp_teacher}

\begin{table*}[t]
\caption{
\textbf{Main results on sequence generation (generative reasoning).}
We evaluate \ours with \qwenfourb and \qweneightb across varying inference budgets.
\textbf{Baselines:} We compare against \textbf{VS} (Verbalized Sampling), \textbf{MT} (Multi-turn), and \textbf{PS} (Parallel Sampling).
Notably, \textbf{CoT} serves as the computational \emph{lower bound} (LB, 1 call), while \textbf{PS (UB)} ($N{=}100$) represents the performance \emph{upper bound}.
\textbf{Metrics:} We report Diversity (\textbf{Div}, $\uparrow$), \textbf{Pass@3} ($\uparrow$), \textbf{ECE$_{\text{top1}}$} and \textbf{ECE$_{\text{cw}}$} ($\downarrow$), and \textbf{NLL} ($\downarrow$).
\ours consistently achieves superior calibration and diversity compared to other low-budget methods.
``$\uparrow$''/``$\downarrow$'' indicate higher/lower is better.
\textbf{Bold} denotes the best result among comparable budgets ($\le 3$ calls).
}

\vspace{-0.6em}
\begin{center}
\resizebox{1\linewidth}{!}{%
\setlength{\tabcolsep}{6pt}
\begin{tabular}{lc ccccc ccccc}
\toprule[0.12em]
\multirow{2}{*}[-0.25em]{\textbf{Method}}
& \multirow{2}{*}[-0.25em]{\textbf{\#Calls}}
& \multicolumn{5}{c}{\textbf{MATH-500}}
& \multicolumn{5}{c}{\textbf{AIME 24\&25}} \\
\cmidrule(lr){3-7} \cmidrule(lr){8-12}
& &
\emph{Div ($\uparrow$)}
& \emph{Pass@3 ($\uparrow$)}
& \emph{ECE$_{\text{top1}}$ ($\downarrow$)}
& \emph{ECE$_{\text{cw}}$ ($\downarrow$)}
& \emph{NLL ($\downarrow$)}
&
\emph{Div ($\uparrow$)}
& \emph{Pass@3 ($\uparrow$)}
& \emph{ECE$_{\text{top1}}$ ($\downarrow$)}
& \emph{ECE$_{\text{cw}}$ ($\downarrow$)}
& \emph{NLL ($\downarrow$)} \\
\midrule

\multicolumn{12}{l}{\qwenfourb} \\
\midrule
\quad CoT (LB)
& 1
& - & \textbf{83.60} & 16.40 & 16.40 & 3.02
& - & 16.67 & 83.33 & 83.33 & 15.35 \\

\quad VS
& 1
& 39.00 & 57.20 & 39.92 & 35.32 & 7.97
& 31.67 & 8.33 & 87.46 & 90.53 & 16.89 \\

\rowcolor{lightergray}
\quad \ours (Ours)
& 1
& \textbf{51.60} & 79.80 & \textbf{8.97} & \textbf{7.63} & \textbf{2.47} 
& \textbf{76.67} & \textbf{18.33} & \textbf{40.21} & \textbf{38.58} & \textbf{11.37} \\

\cmidrule(lr){1-12}

\quad MT
& 3
& 36.73 & 84.20 & 13.80 & 12.97 & 2.85
& 45.00 & 20.00 & 67.22 & 55.81 & 14.75 \\

\quad PS
& 3
& 40.40 & \textbf{84.40} & 9.61 & 8.32 & 2.72
& 70.56 & \textbf{23.33} & 43.45 & 42.57 & 13.84 \\

{\quad VS}
& 3
& \textbf{64.73} & 68.00 & 20.86 & 15.54 & 5.97 
& 72.77 & 16.67 & 46.11 & 38.17 & 15.42 \\

\rowcolor{lightergray}
\quad \ours (Ours)
& 3 
& 57.47 & 80.60 & \textbf{8.53} & \textbf{6.81} & \textbf{2.22} 
& \textbf{81.11} & 20.00 & \textbf{39.87} & \textbf{35.41} & \textbf{11.07} \\

\cmidrule(lr){1-12}

\quad PS (UB)
& 100
& - & 89.60 & 3.15 & 1.57 & 0.64
& - & 46.67 & 19.50 & 1.55 & 8.85 \\
\midrule

\multicolumn{12}{l}{\qweneightb} \\
\midrule
\quad CoT (LB)
& 1
& - & \textbf{84.60} & 15.40 & 15.40 & 2.84
& - & \textbf{23.33} & 76.77 & 76.77 & 14.12 \\

\quad VS
& 1
& 40.73 & 78.20 & 20.21 & 17.57 & 4.05
& 55.56 & 18.33 & 74.04 & 48.04 & 15.05 \\

\rowcolor{lightergray}
\quad \ours (Ours)
& 1
& \textbf{61.27} & 82.80 & \textbf{8.13} & \textbf{6.85} & \textbf{2.17}
& \textbf{72.22} & 20.00 & \textbf{28.11} & \textbf{26.53} & \textbf{10.48} \\

\cmidrule(lr){1-12}

\quad MT
& 3
& 44.73 & \textbf{86.60} & 12.73 & 9.32 & 2.62
& 50.00 & 25.00 & 56.67 & 48.39 & 13.85 \\

\quad PS
& 3
& 46.07 & 85.40 & 9.54 & 8.72 & 2.47
& 75.00 & \textbf{26.67} & 34.13 & 31.58 & 12.15 \\

{\quad VS}
& 3
& 56.47 & 82.80 & 9.70 & 7.34 & 2.56
& \textbf{83.33} & 23.33 & 47.87 & 29.33 & 13.60 \\

\rowcolor{lightergray}
\quad \ours (Ours)
& 3 
& \textbf{61.73} & 83.10 & \textbf{7.97} & \textbf{6.02} & \textbf{2.07}
& 77.22 & 20.00 & \textbf{27.96} & \textbf{24.47} & \textbf{10.16} \\

\cmidrule(lr){1-12}

\quad PS (UB)
& 100
& - & 90.00 & 4.31 & 1.40 & 0.67
& - & 48.33 & 11.93 & 1.34 & 8.25 \\
\bottomrule[0.12em]
\end{tabular}
}
\end{center}
\label{tab:main-qwen}
\vspace{-1em}
\end{table*}

The effectiveness of our \ours framework critically depends on the quality of the teacher model’s supervision. While Bayesian LLMs have demonstrated strong theoretical and empirical success in classification settings, constructing equally reliable supervision signals for long-form sequence generation remains relatively under-explored. 
We hypothesize that the empirical distribution induced by multiple reasoning traces, $\hat{\pi}_{\vtheta}(\vy \mid \vx)$, provides a well-calibrated proxy for the underlying posterior. In this section, we empirically validate this hypothesis by examining how the number of sampled reasoning traces influences the reliability of the resulting distribution.

\textbf{Setup.}\quad
We conduct a systematic evaluation using 10k instances sampled from the DeepScaleR~\cite{luo2025deepscaler} dataset. We vary the sample budget $N$ and measure the quality of the resulting empirical distributions over the final answers using ACC, ECE, and NLL. To ensure robustness, we evaluate 8 representative LLMs across the 3 model families, spanning \texttt{Gemma}, \texttt{Llama}, and \texttt{Qwen}, with all reported results averaged over 100 independent sampling runs to minimize variance. \textbf{The results are summarized in \Figref{fig:teacher-samples},} with detailed numerical results reported in \Tabref{tab:main-iau}.

\textbf{Observations.}\quad
Two key observations emerge from this analysis: 
\emph{(i) As the number of sampled reasoning traces increases, most models achieve strong calibration,} with Expected Calibration Error (ECE) consistently dropping below 10\%. This indicates that trace marginalization yields reliable and well-calibrated supervision targets when sufficient samples are available, a phenomenon we refer to as \textbf{\emph{Internal Alignment of Uncertainty~(IAU)}}.
\emph{(ii) Although the evaluated models exhibit substantial variation in their final predictive accuracy, their calibration metrics (both ECE and NLL) converge to comparable levels.} This relative invariance across model families suggests that the quality of trace-induced uncertainty estimates is largely model-agnostic, highlighting the broad applicability and robustness of our framework across diverse LLM backbones.

Given the observations above, we will adopt \qwenfourb and \qweneightb as representative testbeds for the subsequent experiments of sequence generation tasks.


\subsection{\ours on Generative Reasoning}
\label{sec:results_math}

\Tabref{tab:main-qwen} reports the performance of \ours and baselines on MATH-500 and AIME 24\&25 using \qwenfourb and \qweneightb as backbones under varying inference budgets, while \Tabref{tab:token-stats} summarizes their corresponding token usage.

\begin{table}[t]
\centering
\caption{Average token length of model outputs across different methods and datasets. We compare our approach with three baselines using \texttt{Qwen3-4B/8B Non-thinking} models.}
\label{tab:token-stats}
\vspace{0em}
\resizebox{1\linewidth}{!}{%
\setlength{\tabcolsep}{6pt}
\begin{tabular}{lccccc}
\toprule
\multirow{2}{*}{\textbf{Method}} & \multicolumn{2}{c}{\textbf{Math500}} & & \multicolumn{2}{c}{\textbf{\textbf{AIME 24\&25}}} \\
\cmidrule{2-3} \cmidrule{5-6}
& \texttt{Qwen3-4B} & \texttt{Qwen3-8B} & & \texttt{Qwen3-4B} & \texttt{Qwen3-8B} \\
\midrule
CoT  & 754.31 & 825.08 &  & 2319.95 & 2582.02 \\
VS   & 452.46 & 798.56 &  & 1351.47 & 1484.38 \\
MT   & 1943.42 & 2103.95 & & 7629.62 & 8013.33 \\
PS-3  & 2056.70 & 2669.21 & & 7693.97 & 8048.88 \\

\rowcolor{lightergray}
\ours~(Ours) & {1012.64} & {1057.53} & & {1917.33} & {1693.30} \\
\bottomrule
\end{tabular}
}
\vspace{-1.4em}
\end{table}

\begin{table*}[t]
\caption{
    \textbf{Results on sequence classification~(discriminative reasoning),} where \textbf{ACC} and \textbf{ECE} are reported in percentages. 
    \textbf{``SF?''} stands for whether a method is sampling-free during inference, and we use $N{=}10$ samples in all sampling-based baseline methods.
    \ours uses \blob~\cite{wang2024blob} as the teacher model and is trained for 10,000 iterations. 
    ``$\uparrow$'' and ``$\downarrow$'' indicate that higher and lower values are preferred, respectively. 
    \textbf{Boldface} and \underline{underlining} denote the best and the second-best performance, respectively. 
}
\vspace{-1em}
\begin{center}
\resizebox{1\linewidth}{!}{%
\setlength{\tabcolsep}{7pt}
\begin{tabular}{clcccc ccc cccc}
	\toprule[0.12em]
	\multirow{3}{*}[-0.25em]{\textbf{Metric}} 
    & \multirow{3}{*}[-0.25em]{\textbf{Method}} 
    & \multirow{3}{*}[-0.25em]{\textbf{SF?}}
    & \multicolumn{6}{c}{\multirow{2}{*}[-0.25em]{\textbf{In-Distribution Datasets}}} 
    & \multicolumn{4}{c}{\textbf{Out-of-Distribution Datasets} (OBQA$\rightarrow$X)}
     \\
     \cmidrule(lr){10-13}
     
     & & &
     & & & & & 
     & \multicolumn{2}{c}{\emph{Small Shift}}
     & \multicolumn{2}{c}{\emph{Large Shift}}
     \\
     
     \cmidrule(lr){4-9} \cmidrule(lr){10-11} \cmidrule(lr){12-13}
     & & & WG-S %
     & ARC-C%
     & ARC-E%
     & WG-M%
     & OBQA
     & BoolQ%
     & ARC-C%
     & ARC-E%
     & Chem
     & Phy
     \\
     \midrule

     \multirow{9}{*}{ACC~($\uparrow$)} 
     
    & \loramcd
    & \textcolor{red}{\xmark}
    & \underline{78.03\scriptsize{$\pm$0.61}}  
    & 81.64\scriptsize{$\pm$1.79}  
    & 91.37\scriptsize{$\pm$0.38}  
    & 83.18\scriptsize{$\pm$0.84}  
    & 87.20\scriptsize{$\pm$1.02}  
    & 89.93\scriptsize{$\pm$0.16}  
    & 81.42\scriptsize{$\pm$1.38}  
    & 87.27\scriptsize{$\pm$0.84}  
    & 47.92\scriptsize{$\pm$2.25}  
    & \textbf{46.53\scriptsize{$\pm$0.49}}
    \\
     
    & \loraens
    & \textcolor{red}{\xmark} 
    & \textbf{78.82\scriptsize{$\pm$0.52}} 
    & 82.55\scriptsize{$\pm$0.42}  
    & \textbf{91.84\scriptsize{$\pm$0.36}}  
    & \textbf{83.99\scriptsize{$\pm$0.74}}  
    & 87.37\scriptsize{$\pm$0.67}  
    & \textbf{90.50\scriptsize{$\pm$0.14}} 
    & 79.62\scriptsize{$\pm$0.57}  
    & 86.56\scriptsize{$\pm$0.60}  
    & \textbf{49.65\scriptsize{$\pm$3.22}}  
    & 44.44\scriptsize{$\pm$1.96}
    \\

    & \loralap
    & \textcolor{red}\xmark
    & 76.05\scriptsize{$\pm$0.92}
    & 79.95\scriptsize{$\pm$0.42} 
    & 90.73\scriptsize{$\pm$0.08} 
    & 82.83\scriptsize{$\pm$0.85} 
    & 87.90\scriptsize{$\pm$0.20} 
    & 89.36\scriptsize{$\pm$0.52} 
    & 81.08\scriptsize{$\pm$1.20} 
    & 87.21\scriptsize{$\pm$1.20} 
    & \underline{48.26\scriptsize{$\pm$3.93}} 
    & \underline{46.18\scriptsize{$\pm$1.30}}  
    \\

    & \blob
    & \textcolor{red}\xmark
    & 76.45\scriptsize{$\pm$0.37}
    & 82.32\scriptsize{$\pm$1.15} 
    & 91.14\scriptsize{$\pm$0.54}  
    & 82.01\scriptsize{$\pm$0.56}  
    & 87.57\scriptsize{$\pm$0.21} 
    & 89.65\scriptsize{$\pm$0.15} 
    & 79.75\scriptsize{$\pm$0.43}  
    & 87.13\scriptsize{$\pm$0.00}  
    & 42.71\scriptsize{$\pm$3.71} 
    & 44.79\scriptsize{$\pm$6.64}
    \\

    & \tfb
    & \textcolor{red}\xmark
	& 77.81\scriptsize{$\pm$0.36}  
    & \textbf{83.33\scriptsize{$\pm$0.19}}  
    & \underline{91.76\scriptsize{$\pm$0.48}}  
    & 83.81\scriptsize{$\pm$0.39}  
    & 87.80\scriptsize{$\pm$0.16}  
    & \underline{90.11\scriptsize{$\pm$0.28}} 
    & \textbf{82.93\scriptsize{$\pm$1.54}}  
    & \underline{87.64\scriptsize{$\pm$0.51}}  
    & 39.67\scriptsize{$\pm$7.32}  
    & 37.33\scriptsize{$\pm$6.65} 
    \\

    \cmidrule{2-13}

    & \loramle 
    & \textcolor{Emerald}{\cmark}
    & \textbf{77.87\scriptsize{$\pm$0.54}}
    & 81.08\scriptsize{$\pm$0.48}  
    & \textbf{{91.67\scriptsize{$\pm$0.36}}}  
    & 82.30\scriptsize{$\pm$0.53}  
    & 87.90\scriptsize{$\pm$0.87}  
    & 89.58\scriptsize{$\pm$0.26} 
    & \textbf{81.48\scriptsize{$\pm$2.41}}  
    & \textbf{86.83\scriptsize{$\pm$0.87}}  
    & \textbf{45.83\scriptsize{$\pm$0.85}}  
    & \textbf{42.36\scriptsize{$\pm$1.77}} 
    \\

    & \loramap 
    & \textcolor{Emerald}{\cmark}
    & 76.90\scriptsize{$\pm$0.97}  
    & 81.08\scriptsize{$\pm$2.48}  
    & 91.61\scriptsize{$\pm$0.44}  
    & 82.59\scriptsize{$\pm$0.28}  
    & 85.73\scriptsize{$\pm$0.19}  
    & \textbf{{90.09\scriptsize{$\pm$0.28}}}  
    & 79.98\scriptsize{$\pm$0.87}  
    & 86.58\scriptsize{$\pm$0.79}  
    & 43.40\scriptsize{$\pm$4.98}  
    & 38.54\scriptsize{$\pm$3.40}  
    \\

    & \blob-M
    & \textcolor{Emerald}{\cmark}
    & 77.72\scriptsize{$\pm$0.12}  
    & {82.60\scriptsize{$\pm$0.60}}  
    & 91.64\scriptsize{$\pm$0.55}  
    & \underline{83.92\scriptsize{$\pm$0.48}}  
    & \underline{88.00\scriptsize{$\pm$0.80}}  
    & 89.86\scriptsize{$\pm$0.05}   
    & \underline{82.06\scriptsize{$\pm$1.15}}  
    & \textbf{88.54\scriptsize{$\pm$0.31}}  
    & 39.93\scriptsize{$\pm$5.20}  
    & 39.93\scriptsize{$\pm$4.02} \\

    & \cellcolor{lightergray}\ours~(Ours)
    & \cellcolor{lightergray}\textcolor{Emerald}{\cmark}
    & \cellcolor{lightergray}77.58\scriptsize{$\pm$1.39}  
    & \cellcolor{lightergray}\textbf{82.75\scriptsize{$\pm$0.16}}
    & \cellcolor{lightergray}91.37\scriptsize{$\pm$0.33}  
    & \cellcolor{lightergray}\textbf{83.25\scriptsize{$\pm$0.47}}  
    & \cellcolor{lightergray}\textbf{88.17\scriptsize{$\pm$0.67}}  
    & \cellcolor{lightergray}{90.06\scriptsize{$\pm$0.09}}  
    & \cellcolor{lightergray}78.36\scriptsize{$\pm$0.87}  
    & \cellcolor{lightergray}84.76\scriptsize{$\pm$0.55}  
    & \cellcolor{lightergray}39.58\scriptsize{$\pm$1.70}  
    & \cellcolor{lightergray}39.58\scriptsize{$\pm$4.50}  
    \\
     
     \midrule
    
     \multirow{9}{*}{ECE~($\downarrow$)}

    & \loramcd
    & \textcolor{red}\xmark
    & 16.13\scriptsize{$\pm$0.54}  
    & 13.69\scriptsize{$\pm$1.11}  
    & 6.73\scriptsize{$\pm$0.71}  
    & 13.05\scriptsize{$\pm$0.99}  
    & 9.76\scriptsize{$\pm$0.71}  
    & 7.95\scriptsize{$\pm$0.17}  
    & 13.63\scriptsize{$\pm$1.18}  
    & 9.27\scriptsize{$\pm$0.60}  
    & 30.91\scriptsize{$\pm$3.57}  
    & 33.08\scriptsize{$\pm$1.40} 
    \\
     
    & \loraens
    & \textcolor{red}\xmark
    & 14.72\scriptsize{$\pm$0.17}  
    & 13.45\scriptsize{$\pm$1.19} 
    & 6.59\scriptsize{$\pm$0.45} 
    & 11.17\scriptsize{$\pm$0.92}  
    & 8.17\scriptsize{$\pm$0.86}  
    & 7.35\scriptsize{$\pm$0.55} 
    & 11.37\scriptsize{$\pm$1.82} 
    & 7.21\scriptsize{$\pm$1.13}
    & 18.92\scriptsize{$\pm$6.03} 
    & 26.80\scriptsize{$\pm$3.23} \\

    & \loralap
    & \textcolor{red}\xmark 
    & \textbf{4.18\scriptsize{$\pm$0.11}} 
    & 9.26\scriptsize{$\pm$3.08} 
    & 5.27\scriptsize{$\pm$0.51} 
    & \textbf{3.50\scriptsize{$\pm$0.78}} 
    & 8.93\scriptsize{$\pm$0.34} 
    & \textbf{1.93\scriptsize{$\pm$0.22}} 
    & 7.83\scriptsize{$\pm$1.49} 
    & 7.80\scriptsize{$\pm$1.99} 
    & \textbf{14.49\scriptsize{$\pm$0.57}} 
    & \underline{13.17\scriptsize{$\pm$2.14}} 
    \\
     
    & \blob
    & \textcolor{red}\xmark
    & 9.93\scriptsize{$\pm$0.22} 
    & \textbf{5.41\scriptsize{$\pm$1.17}}  
    & \underline{2.70\scriptsize{$\pm$0.87}}  
    & 4.28\scriptsize{$\pm$0.64}  
    & \underline{2.91\scriptsize{$\pm$0.92}}  
    & \underline{2.58\scriptsize{$\pm$0.25}} 
    & \textbf{5.61\scriptsize{$\pm$0.40}}  
    & \textbf{2.48\scriptsize{$\pm$0.43}}  
    & 16.67\scriptsize{$\pm$0.87}  
    & \textbf{12.78\scriptsize{$\pm$4.18}} 
    \\

    & \tfb
    & \textcolor{red}\xmark
	& \underline{8.16\scriptsize{$\pm$0.48}}  
    & \underline{6.48\scriptsize{$\pm$0.36}}  
    & \textbf{2.44\scriptsize{$\pm$0.50}}  
    & \underline{3.83\scriptsize{$\pm$0.43}}  
    & \textbf{2.67\scriptsize{$\pm$0.18}}  
    & 3.10\scriptsize{$\pm$0.59} 
    & 6.69\scriptsize{$\pm$1.63}  
    & \underline{3.61\scriptsize{$\pm$0.87}}  
    & 18.45\scriptsize{$\pm$6.75}  
    & 20.53\scriptsize{$\pm$6.27}
    \\

     \cmidrule{2-13}

    & \loramle 
    & \textcolor{Emerald}\cmark
    & 17.02\scriptsize{$\pm$0.46}  
    & 16.35\scriptsize{$\pm$0.68}  
    & 7.00\scriptsize{$\pm$0.53}  
    & 13.83\scriptsize{$\pm$0.65}  
    & 9.77\scriptsize{$\pm$0.81}  
    & 8.69\scriptsize{$\pm$0.21} 
    & 14.45\scriptsize{$\pm$2.19}  
    & 10.78\scriptsize{$\pm$0.50}  
    & 32.46\scriptsize{$\pm$2.60}  
    & 38.41\scriptsize{$\pm$4.44}  
    \\

    & \loramap 
    & \textcolor{Emerald}\cmark
    & 18.71\scriptsize{$\pm$0.74}
    & 15.77\scriptsize{$\pm$1.60}
    & 6.62\scriptsize{$\pm$0.64}
    & 14.26\scriptsize{$\pm$0.92}
    & 12.19\scriptsize{$\pm$0.55}
    & 8.40\scriptsize{$\pm$0.25}
    & 16.46\scriptsize{$\pm$0.44}
    & 11.36\scriptsize{$\pm$0.58}
    & 34.79\scriptsize{$\pm$3.76}
    & 38.50\scriptsize{$\pm$2.18}
    \\

    & \blob-M
    & \textcolor{Emerald}\cmark 
    & 15.43\scriptsize{$\pm$0.15}  
    & 12.41\scriptsize{$\pm$1.52}  
    & 4.91\scriptsize{$\pm$0.28}  
    & 9.37\scriptsize{$\pm$1.33}  
    & 6.44\scriptsize{$\pm$0.15}  
    & 6.26\scriptsize{$\pm$0.29}   
    & 11.22\scriptsize{$\pm$0.38}  
    & 6.34\scriptsize{$\pm$0.71}  
    & 26.65\scriptsize{$\pm$3.06}  
    & 25.40\scriptsize{$\pm$5.40} \\

    & \cellcolor{lightergray}\ours~(Ours)
    & \cellcolor{lightergray}\textcolor{Emerald}\cmark
    & \cellcolor{lightergray}\textbf{9.24\scriptsize{$\pm$1.16}}
    & \cellcolor{lightergray}\textbf{4.76\scriptsize{$\pm$0.58}}
    & \cellcolor{lightergray}\textbf{4.49\scriptsize{$\pm$0.56}}
    & \cellcolor{lightergray}\textbf{2.37\scriptsize{$\pm$0.69}}
    & \cellcolor{lightergray}\textbf{2.55\scriptsize{$\pm$0.12}}
    & \cellcolor{lightergray}\textbf{2.34\scriptsize{$\pm$0.28}}
    & \cellcolor{lightergray}\textbf{5.51\scriptsize{$\pm$0.24}}
    & \cellcolor{lightergray}\textbf{4.88\scriptsize{$\pm$0.65}}
    & \cellcolor{lightergray}\textbf{13.74\scriptsize{$\pm$2.80}}
    & \cellcolor{lightergray}\textbf{15.66\scriptsize{$\pm$4.60}}
    \\

     \midrule
    
    \multirow{9}{*}{NLL~($\downarrow$)} 

    & \loramcd
    & \textcolor{red}\xmark
    & 0.83\scriptsize{$\pm$0.01}  
    & 0.99\scriptsize{$\pm$0.10}  
    & 0.45\scriptsize{$\pm$0.06}  
    & 0.64\scriptsize{$\pm$0.03}  
    & 0.62\scriptsize{$\pm$0.08}  
    & 0.49\scriptsize{$\pm$0.01} 
    & 1.03\scriptsize{$\pm$0.02} 
    & 0.61\scriptsize{$\pm$0.03}  
    & 1.91\scriptsize{$\pm$0.18}  
    & 2.02\scriptsize{$\pm$0.15} 
    \\
     
    & \loraens
    & \textcolor{red}\xmark
    & 0.75\scriptsize{$\pm$0.02}  
    & 0.80\scriptsize{$\pm$0.11}  
    & 0.38\scriptsize{$\pm$0.03}  
    & 0.55\scriptsize{$\pm$0.02}  
    & 0.45\scriptsize{$\pm$0.05}  
    & 0.42\scriptsize{$\pm$0.05} 
    & 0.72\scriptsize{$\pm$0.07}  
    & \underline{0.44\scriptsize{$\pm$0.03}}  
    & 1.40\scriptsize{$\pm$0.18}  
    & 1.50\scriptsize{$\pm$0.13} 
    \\

    & \loralap
    & \textcolor{red}\xmark
    & \underline{0.56\scriptsize{$\pm$0.00}} 
    & 1.18\scriptsize{$\pm$0.02} 
    & 1.04\scriptsize{$\pm$0.01} 
    & 0.51\scriptsize{$\pm$0.00} 
    & 0.94\scriptsize{$\pm$0.00} 
    & 0.43\scriptsize{$\pm$0.00} 
    & 1.17\scriptsize{$\pm$0.01} 
    & 1.11\scriptsize{$\pm$0.00} 
    & \textbf{1.27\scriptsize{$\pm$0.01}} 
    & \textbf{1.28\scriptsize{$\pm$0.00}} 
    \\

    & \blob
    & \textcolor{red}\xmark 
    & 0.58\scriptsize{$\pm$0.00}  
    & \textbf{0.51\scriptsize{$\pm$0.03}}  
    & \textbf{0.23\scriptsize{$\pm$0.01}}  
    & \underline{0.43\scriptsize{$\pm$0.01}}  
    &  \underline{0.34\scriptsize{$\pm$0.01}}  
    & \textbf{0.26\scriptsize{$\pm$0.01}}  
    & \underline{0.56\scriptsize{$\pm$0.02}}  
    & \textbf{0.35\scriptsize{$\pm$0.02}}  
    & \underline{1.34\scriptsize{$\pm$0.04}}  
    & \underline{1.35\scriptsize{$\pm$0.10}} 
    \\

    & \tfb
    & \textcolor{red}\xmark
    & \textbf{0.55\scriptsize{$\pm$0.01}}  
    & \underline{0.53\scriptsize{$\pm$0.04}}  
    & \textbf{0.23\scriptsize{$\pm$0.02}}  
    & \textbf{0.40\scriptsize{$\pm$0.01}}  
    & \textbf{0.33\scriptsize{$\pm$0.02}}  
    & \underline{0.27\scriptsize{$\pm$0.01}} 
    & \textbf{0.52\scriptsize{$\pm$0.05}}  
    & \textbf{0.35\scriptsize{$\pm$0.02}}  
    & 1.36\scriptsize{$\pm$0.13}  
    & 1.46\scriptsize{$\pm$0.11} 
    \\

    \cmidrule{2-13}

    & \loramle 
    & \textcolor{Emerald}\cmark
    & 0.88\scriptsize{$\pm$0.04}  
    & 1.20\scriptsize{$\pm$0.11}  
    & 0.46\scriptsize{$\pm$0.04}  
    & 0.68\scriptsize{$\pm$0.01}  
    & 0.61\scriptsize{$\pm$0.06}  
    & 0.52\scriptsize{$\pm$0.01}   
    & 1.07\scriptsize{$\pm$0.06}  
    & 0.72\scriptsize{$\pm$0.06}  
    & 1.91\scriptsize{$\pm$0.16}  
    & 2.25\scriptsize{$\pm$0.21}  
    \\

    & \loramap 
    & \textcolor{Emerald}\cmark
    & 0.99\scriptsize{$\pm$0.07} 
    & 1.12\scriptsize{$\pm$0.23} 
    & 0.46\scriptsize{$\pm$0.03} 
    & 0.74\scriptsize{$\pm$0.07} 
    & 0.79\scriptsize{$\pm$0.02} 
    & 0.52\scriptsize{$\pm$0.01} 
    & 1.19\scriptsize{$\pm$0.04} 
    & 0.83\scriptsize{$\pm$0.06}  
    & 1.97\scriptsize{$\pm$0.13} 
    & 2.32\scriptsize{$\pm$0.10} 
    \\

    & \blob-M
    & \textcolor{Emerald}\cmark
    & 0.74\scriptsize{$\pm$0.02}  
    & 0.73\scriptsize{$\pm$0.04}  
    & \underline{0.29\scriptsize{$\pm$0.03}}  
    & 0.47\scriptsize{$\pm$0.03}  
    & 0.37\scriptsize{$\pm$0.02}  
    & 0.32\scriptsize{$\pm$0.02}   
    & 0.67\scriptsize{$\pm$0.07}  
    & 0.39\scriptsize{$\pm$0.03}  
    & 1.53\scriptsize{$\pm$0.13}  
    & 1.54\scriptsize{$\pm$0.15} 
    \\

    & \cellcolor{lightergray}\ours~(Ours)
    & \cellcolor{lightergray}\textcolor{Emerald}\cmark
    & \cellcolor{lightergray}\textbf{0.55\scriptsize{$\pm$0.03}} 
    & \cellcolor{lightergray}\textbf{0.52\scriptsize{$\pm$0.03}} 
    & \cellcolor{lightergray}\textbf{0.24\scriptsize{$\pm$0.01}} 
    & \cellcolor{lightergray}\textbf{0.41\scriptsize{$\pm$0.02}} 
    & \cellcolor{lightergray}\textbf{0.33\scriptsize{$\pm$0.02}} 
    & \cellcolor{lightergray}\textbf{0.24\scriptsize{$\pm$0.00}} 
    & \cellcolor{lightergray}\textbf{0.61\scriptsize{$\pm$0.01}} 
    & \cellcolor{lightergray}\textbf{0.40\scriptsize{$\pm$0.00}}  
    & \cellcolor{lightergray}\textbf{1.32\scriptsize{$\pm$0.06}} 
    & \cellcolor{lightergray}\textbf{1.34\scriptsize{$\pm$0.02}} 
    \\

    \bottomrule[0.12em]
    \end{tabular}
 }
\end{center}
\label{tab:main-llama}
\vspace{-1em}
\end{table*}

As shown in \Tabref{tab:main-qwen}, among methods restricted to a single forward pass, \ours achieves \textbf{substantially better calibration while largely preserving accuracy}. In the strict one-pass setting, \textbf{VS} provides explicit probability estimates but often severely degrades reasoning performance. For example, on MATH-500 with \qwenfourb, Pass@3 drops from 83.60 to 57.00. In contrast, \ours attains significantly lower ECE (approximately $2{\times}$ improvement), lower NLL, and higher diversity, while maintaining competitive Pass@3. This demonstrates that \ours improves uncertainty quality without the severe accuracy loss typically incurred when producing multiple reasoning traces within a single forward pass.


Moreover, \ours’s \textbf{uncertainty estimation capability scales favorably with test-time compute}. Increasing the inference budget from one to three calls consistently improves both calibration (ECE/NLL) and Pass@3, indicating that distributional self-distillation is complementary to test-time sampling rather than a substitute for it.
\ours also \textbf{maintains strong performance on more challenging benchmarks}. When moving from MATH-500 to AIME 24\&25, baseline methods exhibit pronounced degradation in ECE and NLL, whereas \ours sustains low NLL with stable ECE. This suggests that distributional supervision enhances robustness under increased task difficulty and distribution shift by capturing intrinsic uncertainty in complex reasoning.

Finally, \Tabref{tab:token-stats} shows that \ours \textbf{produces markedly shorter outputs than multi-sample baselines}, including Multi-Turn~(MT) and Parallel Sampling with three samples (PS-3), across both datasets, thereby reducing decoding overhead. Notably, while Verbalized Sampling~(VS) significantly reduces output length relative to its CoT baseline, it does so at the cost of reasoning quality, indicating that models without \ours-style training struggle to reliably produce distributional responses. Taken together with \Tabref{tab:main-qwen}, these results demonstrate that \ours achieves a more favorable reliability–compute trade-off.


\subsection{\ours on Discriminative Reasoning}
\label{sec:results_commonsense}

\Tabref{tab:main-llama} summarizes the performance on commonsense reasoning benchmarks using \texttt{Llama-3.1-8B}. 

Across six \emph{in-distribution} benchmarks, \ours \textbf{closely matches the teacher in both accuracy and uncertainty quality}. In terms of accuracy, \ours is competitive with BLoB, achieving slight gains on OBQA (88.17 vs. 87.57) and BoolQ (90.06 vs. 89.65), with only minor regressions on a few tasks (e.g., ARC-E: 89.52 vs. 91.14). Overall, accuracy is well preserved and consistently outperforms other sampling-free baselines (MLE, MAP), while remaining competitive with sampling-based methods.
Crucially, \ours provides strong calibration: it substantially reduces ECE (up to $2{\times}$) and improves NLL by as much as 0.1. Across all in-distribution datasets, \ours \textbf{achieves the best ECE and NLL}, demonstrating consistently reliable uncertainty estimation compared to prior approaches.


We evaluate uncertainty generalization by fine-tuning on OBQA and testing \emph{under distributional shifts}.
Under \emph{small distributional shifts}, \ours remains robust: accuracy reaches 78.36 (ARC-C) and 84.76 (ARC-E), close to OBQA performance and slightly below BLoB. Importantly, \ours maintains strong calibration, achieving ECE of 5.51 (ARC-C) and 4.88 (ARC-E), outperforming other sampling-free baselines and remaining competitive with BLoB. 
Under \emph{large distributional shifts}, all methods experience substantial accuracy drops; nevertheless, \ours retains robust uncertainty estimation. Accuracy is comparable to BLoB-M, while calibration improves: ECE on Chemistry drops to 13.74 (vs. 16.67 for BLoB), outperforming sampling-based baselines such as TFB and ENS. Notably, \ours achieves the second-best NLL overall under large shifts and remains the strongest by a wide margin among sampling-free methods, indicating that distributional distillation yields transferable uncertainty under domain shift.

\textbf{Remark on Efficiency.}\quad
\ours is sampling-free (SF?=\textcolor{Emerald}{\cmark}) in sequence classification tasks, and requires only a single forward pass at test time.
This contrasts with sampling-based baselines that use $N{=}10$ samples and strictly improves the test-time efficiency by 10 times.

\section{Conclusion}
\label{sec:conclusion}

In this work, we propose \ours, a unified distributional distillation framework that compresses trace/weight-based uncertainty into a single inference pass of LLMs. 
Particularly, quite contrary to the prevailing view that LLMs are inherently miscalibrated, we identify \emph{Internal Alignment of Uncertainty (IAU)}, showing that predictive distributions induced by reasoning traces are intrinsically well calibrated.
Extensive experiments across generative reasoning and discriminative tasks demonstrate that \ours achieves state-of-the-art uncertainty estimation capability while preserving accuracy. By bridging efficiency and reliability, our method matches uncertainty quality with lower inference cost, highlighting distributional distillation as a promising direction for post-training trustworthy LLMs.

\section*{Impact Statement}
This research advances methods for making large language models more capable, reliable, and interpretable, through distributional distillation that makes one-pass uncertainty estimation possible. There are many potential societal consequences of our work, none of which we feel must be specifically highlighted here.


\bibliographystyle{icml2026}
\bibliography{ref,ref_bayesian_llms,ref_llm_reasoning}

\onecolumn
\appendix
\clearpage
\section*{\LARGE Appendix}
\markboth{Appendix}{Appendix}

In Appendix~\ref{app:algorithm}, we present the full algorithmic description of \ours, including the curriculum schedule details. In Appendix~\ref{app:limitations}, we discuss the limitations of our current framework. Next, in Appendix~\ref{app:exp_settings}, we provide detailed experimental settings, covering evaluation metrics, baseline descriptions, details, and prompt templates. Finally, in Appendix~\ref{app:more}, we present additional empirical results, including:
\begin{itemize}[nosep]
    \item For \emph{sequence classification} tasks~(\appref{app:additional-results-sc})
    \begin{itemize}[nosep]
        \item an ablation study on the impact of different distillation losses (\appref{subsubsec:ablation_distillation_loss}),
        \item the rationale behind our simple loss scheduling strategy (\appref{app:simple_loss_s}),
        \item the impact of dataset augmentation on student model performance (\appref{app:data_aug}),
        \item a performance analysis of off-the-shelf LLMs' internal predictive distributions (\appref{app:internal_dist}),
        \item additional downstream adaptation results on the Llama3.1-8B architecture (\appref{app:llama_lora_more}).
    \end{itemize}
    \item For \emph{sequence generation} tasks~(\appref{app_details_sg})
    \begin{itemize}[nosep]
        \item the details of our trace cleaning pipeline for data construction (\appref{app:details_1}),
        \item a qualitative comparison between \ours and baseline methods (\appref{app:comparsion}).
    \end{itemize}
    
\end{itemize}


\section{Algorithm}
\label{app:algorithm}

\begin{algorithm}[h]
\caption{
\textbf{Dist}ributional \textbf{Distill}ation {(\ours)} for Sequence Generation
}
\label{alg:ours}
\begin{algorithmic}[1]
\REQUIRE
$\gD_{\text{train}}$: Training query dataset. \par
$\pi_{\vtheta}(\cdot\mid\vx)$: Current model used for self-sampling. \par
$N$: Number of sampled traces per query; $K$: Top-$K$ truncation size. \par
$\tau$: Special delimiter token; confidence head $\hat{\vp}=\mathrm{softmax}(\mathrm{MLP}(\vh_{\tau}))\in\Delta^{K+1}$. \par
$\mathrm{Canon}(\cdot)$: Answer canonicalization (applied to $\vy$). \par
$\eta$: Learning rate; $B_s$: Batch size; $T_{\max}$: Max training steps. \par
$\lambda_t$: Curriculum schedule weight (\appref{app:imple_details}). \par
\ENSURE Trained parameters $\vtheta$.

\STATE \textbf{Phase 1: Offline Data Construction (Distributional Responses)}
\FOR{each $\vx \in \gD_{\text{train}}$}
    \STATE Sample $N$ traces and answers $\{(\vz_n,\vy_n)\}_{n=1}^{N} \overset{\text{i.i.d.}}{\sim} \pi_{\vtheta}(\cdot\mid\vx)$. \hfill$\rhd$ Self-sampling.
    \STATE Canonicalize answers: $\tilde{\vy}_n \leftarrow \mathrm{Canon}(\vy_n)$ for all $n$. \hfill$\rhd$ Assume canonicalized in notation.
    \STATE Compute empirical distribution
    $\hat{\pi}_{\vtheta}(\vy\mid\vx)\leftarrow \frac{1}{N}\sum_{n=1}^{N}\mathbbm{1}_{[\tilde{\vy}_n=\vy]}$. \hfill$\rhd$ \Eqref{eq:trace-induced}.
    \STATE Let $\{(\vy_k,\pi_k)\}_{k\in[K]}$ be the $K$ most frequent answers ordered by $\pi_k$,
    where $\pi_k\triangleq \hat{\pi}_{\vtheta}(\vy_k\mid\vx)$. \hfill$\rhd$ Top-$K$ truncation.
    \STATE Set $(\vy_{K+1},\pi_{K+1}) \leftarrow (\texttt{``others''},\, 1-\sum_{k=1}^{K}\pi_k)$. \hfill$\rhd$ Aggregate tail mass.
    \FOR{$k=1,\cdots,K$}
        \STATE Sample $\vz_k' \sim \hat{\pi}_{\vtheta}(\vz\mid\vy_k,\vx)$ uniformly from
        $\{\vz_n \mid \tilde{\vy}_n=\vy_k\}_{n\in[N]}$. \hfill$\rhd$ Empirical posterior over traces.
    \ENDFOR
    \STATE Form $\mS \leftarrow \{(\vz_k',\vy_k,\pi_k)\}_{k\in[K+1]}$. \hfill$\rhd$ Triplets in the text.
    \STATE Construct SFT target
    $\vs(\vx,\mS)\leftarrow \bigoplus_{k=1}^{K+1}\bigl(\vz_k' \oplus \vy_k \oplus \tau\bigr)$. \hfill$\rhd$ Definition of $\vs(\vx,\mS)$.
\ENDFOR
\STATE $\hat{\gD}_{\text{train}} \leftarrow \{(\vx,\mS,\vs(\vx,\mS),\vpi(\mS))\}$ where $\vpi(\mS)=[\pi_1,\ldots,\pi_{K+1}]^\top$. \par

\STATE \textbf{Phase 2: Training Objective}
\FOR{$t=1,\cdots,T_{\max}$}
    \STATE Sample mini-batch $(\mX_t,\mS_t)\sim \hat{\gD}_{\text{train}}$. \hfill$\rhd$ Batch.
    \STATE Compute SFT loss
    $\mathcal{L}_{\text{sft}} \leftarrow \E\!\left[-\log \pi_{\vtheta}\!\left(\vs(\mX_t,\mS_t)\mid \mX_t\right)\right]$. \hfill$\rhd$ \Eqref{eq:Lsft}.
    \STATE Predict confidence distribution $\hat{\vp}\leftarrow \pi_{\vtheta}(\vy\mid \mX_t,\mS_t)$
    using the delimiter token $\tau$ in $\vs(\mX_t,\mS_t)$. \hfill$\rhd$ MLP head at $\tau$.
    \STATE Compute calibration loss
    $\mathcal{L}_{\text{ce}} \leftarrow \E\!\left[-\sum_{k=1}^{K+1}\pi_k \log \hat{p}_k\right]$. \hfill$\rhd$ \Eqref{eq:Lce}.
    \STATE Total objective
    $\mathcal{L}_{\text{gen}} \leftarrow \mathcal{L}_{\text{sft}} + \lambda_t \mathcal{L}_{\text{ce}}$. \hfill$\rhd$ Definition of $\gL_{\text{gen}}$.
    \STATE Update parameters
    $\vtheta \leftarrow \vtheta - \eta \frac{\partial \mathcal{L}_{\text{gen}}}{\partial \vtheta}$. \hfill$\rhd$ Gradient update.
\ENDFOR
\end{algorithmic}
\end{algorithm}

\begin{algorithm}[h]
\caption{
    \textbf{Dist}ributional \textbf{Distill}ation {(\ours)} for Sequence Classification
}
    \label{alg:eud}
\begin{algorithmic}[1]
\REQUIRE
$\gD_{\text{train}} = \{(\vx_i, \vy_i)\}_{i=1}^M$: Training dataset. \par
$q(\cdot\mid\vphi)$: Bayesian teacher model's approximate weight posterior. \par
$N$: Number of samples for estimation of the teacher model's prediction. \par
$\vtheta_S^{(0)}$: Initial student model weights. \par
$\eta$: Learning rate; $B_s$: Batch size; $T_{\text{max}}$: Number of max training steps. \par
$\mathcal{L}_{\text{KD}}$: Distillation loss function.
\hfill$\rhd$ E.g., KL as in \Eqref{eq:kl-div}. \par
$\alpha_{\text{init}}, \alpha_{\text{final}}$: Initial and final weights for distillation loss. \par
$T_{\alpha}$: Number of annealing steps for $\alpha_t$. \par

\FOR{$i=1,\cdots,M$}
    \STATE $p_T(\vy\mid\vx_i) \leftarrow \frac{1}{N}\sum\nolimits_{n=1}^N \pi_{\vtheta_n}(\vy\mid\vx_i), \quad\vtheta_n\sim q(\cdot\mid\vphi)$. \hfill$\rhd$ Construct teacher model's predictive distributions. 
\ENDFOR
\STATE $\hat{\gD}_{\text{train}}\leftarrow \{(\vx_i, \vy_i, p_T(\vy\mid\vx_i))\}_{i=1}^M$.

\STATE $\vtheta_S \leftarrow \vtheta_S^{(0)}$. \hfill$\rhd$ Student model initialization.
\FOR{$t=1,\cdots,T$}
    \STATE $(\mX_t, \mY_t, \mP_t)\sim \hat{\gD}_{\text{train}}$. \hfill$\rhd$ Sample a mini-batch of data.

    \STATE $\alpha_t \leftarrow \min(\alpha_{\text{init}} + t(\frac{\alpha_{\text{final}}-\alpha_{\text{init}}}{T_\alpha}), \alpha_{\text{final}})$. \hfill$\rhd$ Distillation loss scheduling.
    \STATE $\hat{\mP}_{t} \leftarrow p(\vy\mid\mX_t, \vtheta_S)$. \hfill$\rhd$ Student model's predictive distribution.
        
    \STATE $\gL_t \leftarrow \alpha_t \mathcal{L}_{\text{KD}}(\hat{\mP}_t, \mP_t) + (1-\alpha_{t}) \gL_{\text{CE}}(\hat{\mP}_t, \mY_t)$. \hfill$\rhd$ \Eqref{eq:total}.

    \STATE $\vtheta_S \leftarrow \vtheta_S - \eta \frac{\partial \gL_t}{\partial \vtheta_S}$. \hfill$\rhd$ Gradient update.
        
\ENDFOR
\RETURN Trained student parameters $\vtheta_S$.
\end{algorithmic}
\end{algorithm}


\section{Limitations}
\label{app:limitations}

While \ours demonstrates significant advantages, there are remaining challenges that merit future investigation. First, our framework fundamentally relies on the quality of the teacher's trace-induced distribution; for extremely difficult queries where the base model fails to sample any valid reasoning paths, the student model may struggle to recover a correct probability assignment, suggesting a need for integrating external verification or weak-to-strong generalization techniques. Second, the current approach employs a static structured template with a fixed number of candidates, which may not be optimal for all query types in terms of efficiency and flexibility. A promising direction for future research is to develop an adaptive decoding mechanism that allows the model to autonomously determine the optimal number of candidates to generate based on real-time uncertainty estimates, thereby enabling more flexible and adaptive computation.

\section{Implementation Details}
\label{app:exp_settings}

\subsection{Evaluation Metrics.}
\label{app:exp_settings_metrics}

Let $\mathcal{D}=\{(\vx_i, \vy_i^*)\}_{i=1}^N$ denote the evaluation set, where $\vx_i$ is the input sequence and $\vy_i^*$ is the ground-truth answer sequence.
For each input $\vx_i$, the model outputs $K$ candidate answer--probability pairs $\{(\vy_{i,k}, p_{i,k})\}_{k=1}^K$.

\textbf{Canonicalization.}\quad 
We apply a task-specific canonicalization function $\mathrm{canon}(\cdot)$ to both predictions and ground truth before evaluation.
Define $\tilde{\vy}_{i,k}=\mathrm{canon}(\vy_{i,k})$ and $\tilde{\vy}_i^*=\mathrm{canon}(\vy_i^*)$.
For notational simplicity, in what follows we write $\vy_{i,k}$ and $\vy_i^*$ to refer to these canonicalized answers.

We define a binary correctness indicator $r_{i,k}$ that is 1 if $\vy_{i,k}=\vy_i^*$, and 0 otherwise.

\textbf{Diversity Score (Div).}\quad
To quantify the exploration capability within the one-call budget, we compute the reasoning diversity.
For each problem $\vx_i$, let $U_i$ be the number of unique canonicalized answers in the generated set $\{\vy_{i,1}, \dots, \vy_{i,K}\}$.
The Diversity Score is the average ratio of unique answers:
\begin{equation}
\label{eq:diversity}
\text{Div} = \frac{1}{N} \sum\nolimits_{i=1}^N \frac{U_i}{K}.
\end{equation}

A higher score indicates that the model is effectively exploring the solution space rather than collapsing into a single mode.

\textbf{Calibration Metric (ECE).}\quad 
We evaluate calibration using Expected Calibration Error with $B$ disjoint bins $\{\mathcal{B}_m\}_{m=1}^B$.

\emph{Top-1 ECE.}\quad
Let $k^*_i = \arg\max_k p_{i,k}$ be the index of the top candidate.
Defining $D_m = \{ i \in [N] \mid p_{i, k^*_i} \in \mathcal{B}_m \}$, we compute:
\begin{equation}
\label{eq:ece-top1}
\text{ECE}_{\text{top1}} = \sum\nolimits_{m=1}^B \frac{1}{N} \bigg| \sum\nolimits_{i \in D_m} (r_{i, k^*_i} - p_{i, k^*_i}) \bigg|.
\end{equation}

\emph{Class-wise ECE.}\quad
We measure calibration across all $K$ slots.
Let $D_{m,k} = \{ i \in [N] \mid p_{i,k} \in \mathcal{B}_m \}$.
The metric is averaged over all slots:
\begin{equation}
\label{eq:ece-cw}
\text{ECE}_{\text{cw}} = \frac{1}{N K} \sum\nolimits_{k=1}^K \sum\nolimits_{m=1}^B \bigg| \sum\nolimits_{i \in D_{m,k}} (r_{i,k} - p_{i,k}) \bigg|.
\end{equation}

\textbf{Negative Log-Likelihood (NLL).}\quad
We use NLL as a proper scoring rule.
Defining $P_i(\text{true}) = \sum_{k=1}^K p_{i,k} \cdot r_{i,k}$, the NLL is:
\begin{equation}
\label{eq:nll}
\text{NLL} = - \frac{1}{N} \sum\nolimits_{i=1}^N \log \left( \sum\nolimits_{k=1}^K p_{i,k} \cdot r_{i,k} + \epsilon \right).
\end{equation}

\begin{figure}[t]
    \centering
    \includegraphics[width=.8\linewidth]{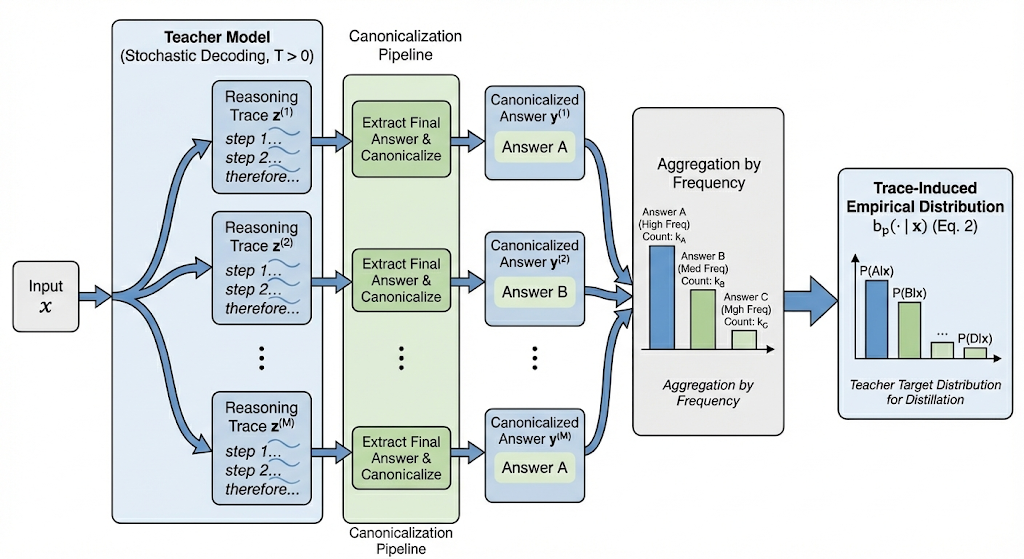} 
    \caption{Offline distribution construction pipeline for \ours.}
    \label{fig:offline_dist}
\end{figure}

\subsection{Baselines}
\label{app:baseline_details}

\textbf{Sequence Generation Baselines.}\quad
We compare \ours against a comprehensive set of training-free and sampling-based approaches:
\begin{itemize}[nosep]
\item \textbf{Chain-of-Thought (CoT)}~\cite{wei2022chain}: Standard greedy decoding (temperature $T=0$) with a single inference pass. This serves as the \emph{computational lower bound (LB)} and baseline for single-solution accuracy. 
\item \textbf{Verbalized Sampling (VS)}~\cite{zhang2025verbalized}: A strong one-call baseline where the model is prompted to generate multiple options and verbally assign probabilities to them in a single pass. \item \textbf{Multi-turn Sampling (MT)}: An iterative sampling approach where the model generates $K$ candidate solutions sequentially, conditioned on the history of previous outputs to refine answers. 
\item \textbf{Parallel Sampling (PS)}~\cite{brown2024large}: Standard Best-of-$N$ majority voting where $N$ independent reasoning paths are sampled. We report results for $N=3$ (denoted as \textbf{PS}) and $N=100$ (denoted as \textbf{PS (UB)}).
\end{itemize}

\textbf{Sequence Classification Baselines.}\quad
We compare \ours with state-of-the-art Bayesian LLMs with LoRA adapters, including 
ensemble-based method: Deep Ensemble (\textbf{\loraens})~\cite{lakshminarayanan2017simple,balabanov2024uncertainty,wang2023lora}, 
variational inference methods: Monte-Carlo Dropout (\textbf{\loramcd})~\cite{gal2016dropout}, 
Bayesian LoRA by Backprop~(\textbf{\blob})~\cite{wang2024blob}, and post-training method: Laplace-LoRA~(\textbf{\loralap})~\cite{yang2023bayesian} and Training-Free Bayesianization~(\textbf{\tfb})~\cite{shi2024training}.
We also include three non-Bayesian methods~(point estimation of the model weights), including two standard PEFT baselines: Maximum Likelihood Estimation~(\textbf{MLE})~\cite{hu2022lora} and Maximum A Posteriori~(\textbf{MAP}), and the weight mean of \blob (\textbf{\blob-M}).
All baselines are implemented following the protocols established in \blob and \tfb~\cite{wang2024blob,shi2024training}.

\subsection{Implementation Details of \ours}
\label{app:imple_details}
\ours is implemented as a two-stage pipeline. 
\textbf{(i) Offline distribution construction:} the predictive distributions are either constructed by sampling from a Bayesian teacher model (classification) or from itself (generation) on the training data; 
\textbf{(ii) Distributional distillation:} using these predictive distributions as targets for our uncertainty distillation. 
We conduct all our experiments on an NVIDIA A100 GPU server.

\textbf{For Sequence Classification,} we use BLoB~\cite{wang2024blob} as the Bayesian teacher model and sample $N{=}100$ times for estimating its predictive distributions, as a larger number of samples during inference improves uncertainty estimation~\cite{wang2024blob}. To maintain parameter efficiency during both training and testing, the student model is implemented using LoRA~\cite{hu2022lora} with rank $r{=}8$ (matching the baselines). We optimize \ours with AdamW~\cite{loshchilov2017decoupled} with $\eta{=}2.75{\times}10^{-4}$, batch size 16, for 10,000 steps. 
    
\textbf{For Sequence Generation,} for each training query $\vx$ in the subset of DeepScaleR~\cite{luo2025deepscaler}, we sample $K{=}3$ teacher reasoning traces using stochastic decoding (temperature $T{=}0.7$, top-$p{=}0.95$). We then apply a GPT-based cleaning step to standardize the raw traces and filter malformed outputs. After cleaning, we extract final answers from the traces and aggregate canonicalized answers by frequency to form an empirical predictive distribution over answers(show in \Figref{fig:offline_dist}). We then train \texttt{Qwen3-4B/8B} with AdamW ($\eta{=}2{\times}10^{-5}$, cosine decay), global batch size 64, for 2 epochs, and warm up the probability-head loss weight $\lambda_t$ over the first 3\% of steps.

Additionally, we employ \emph{curriculum schedule for $\lambda_t$.}
Learning a rigid structured format and a calibrated probability allocation simultaneously can present optimization instabilities during early post-training. We thus employ a curriculum that prioritizes content and format acquisition before enforcing distributional alignment. Specifically, we anneal the weight $\lambda_t$ via a delayed linear ramp:
\begin{equation}
\lambda_t = \lambda_{\max} \cdot \min\left(1, \max\left(0, \tfrac{t-T_0}{T_\lambda}\right)\right),
\label{eq:lambda_schedule}
\end{equation}
where $T_0$ is a burn-in period for stable generation, and $T_\lambda$ controls the transition to distribution matching. This schedule ensures stable convergence while aligning the student’s probability head with the teacher distribution.

\subsection{Prompt Templates}
\label{app:prompts}
This subsection presents the prompt templates we used.
\begin{tcolorbox}[
    breakable,
    enhanced,
    left=-0.5cm, right=-0.5cm, top=2pt, bottom=2pt,
    enlarge top by=0.1cm,
    enlarge bottom by=0.1cm,
    title={\hspace{0.5cm} Chain-of-Thought},
    fonttitle=\bfseries\small
]
\begin{quote}\small
You are a precise math problem solver. Solve the given math problem step by step.\\

\textbf{Prompt:}\\
\texttt{QUESTION: \{question\}}\\
The last line of your response must be exactly of the following format:\\
\texttt{"Therefore, the final answer is: boxed \{ANSWER\}."}
\end{quote}
\end{tcolorbox}

\begin{tcolorbox}[
    breakable,
    enhanced,
    left=-0.5cm, right=-0.5cm, top=2pt, bottom=2pt,
    enlarge top by=0.1cm,
    enlarge bottom by=0.1cm,
    title={\hspace{0.5cm} Trace Cleaning},
    fonttitle=\bfseries\small
]
\begin{quote}\small
You are a data cleaning assistant.\\

\textbf{Prompt:}\\
\texttt{
Your task is to rewrite the solution to be shorter and cleaner,
while preserving the ORIGINAL solution’s structure, reasoning flow,
and writing style as much as possible.
}\\

\texttt{
IMPORTANT PRINCIPLES:\\
- This is NOT style normalization.\\
- This is NOT reformatting to a fixed template.\\
- The cleaned solution should look like the SAME author wrote it, just more concisely.
}

\texttt{
STRICT REQUIREMENTS:\\
1) Faithfulness:\\
   - Do NOT change the meaning.\\
   - Do NOT change the final answer(s).\\
2) Reasoning preservation:\\
   - Keep ALL logical reasoning steps and ALL necessary mathematical derivations.\\
   - You may remove repeated steps, detours, or restatements ONLY if the reasoning remains intact.\\
3) Language \& style preservation:\\
   - Preserve the original narrative tone (e.g. exploratory, explanatory, corrective).\\
   - Preserve structural choices used in the original solution
     (such as step labels, section headers, or paragraph structure),
     unless they are clearly redundant.\\
4) Formatting:\\
   - Do NOT convert narrative explanations into bullet fragments or terse notes.\\
   - Do NOT collapse structured explanations into formula-only derivations.\\
   - Keep LaTeX for mathematics.\\
5) Final answer rule (VERY IMPORTANT):\\
   - End the response with a line exactly in this format:\\
     Final Answer: \boxed{...}\\
   - If there are multiple final answers:\\
     Final Answer: \boxed{...}, \boxed{...}\\
   - Do NOT box intermediate results.\\
   - The “Final Answer:” line must be the LAST line of the output.\\
}

\texttt{Output ONLY the cleaned solution text. No extra commentary.}
\end{quote}
\end{tcolorbox}

\begin{tcolorbox}[
    breakable,
    enhanced,
    left=-0.5cm, right=-0.5cm, top=2pt, bottom=2pt,
    enlarge top by=0.1cm,
    enlarge bottom by=0.1cm,
    title={\hspace{0.5cm} Verbalized Sampling},
    fonttitle=\bfseries\small
]
\begin{quote}\small
You are a precise math problem solver. Solve the given math problem step by step.\\

\textbf{Prompt:}\\
\texttt{QUESTION: \{question\}}\\
\texttt{For each query, please generate a set of three possible responses, each within a separate <response> tag. Responses should each include a numeric <probability>. The final answer in each response is framed as '\$\textbackslash\textbackslash boxed\{ANSWER\}\$ <probability>probs<\textbackslash\textbackslash probability>'.}
\end{quote}
\end{tcolorbox}

\begin{tcolorbox}[
    breakable,
    enhanced,
    left=-0.5cm, right=-0.5cm, top=2pt, bottom=2pt,
    enlarge top by=0.1cm,
    enlarge bottom by=0.1cm,
    title={\hspace{0.5cm} Multiturn-Sampling},
    fonttitle=\bfseries\small
]
\begin{quote}\small
You are a precise math problem solver. Solve the given math problem step by step.\\

\textbf{Turn-1 prompt:}\\
\texttt{QUESTION: \{question\}}\\
The last line of your response must be exactly of the following format:\\
\texttt{"Therefore, the final answer is: \textbackslash\textbackslash boxed\{ANSWER\}."}

\medskip
\textbf{Turn-2 prompt:}\\
\texttt{[Turn-1-prompt] + [Turn-1-Response]}\\
\texttt{"Please provide a completely different final answer from your previous response, while keeping the same output format."}

\medskip
\textbf{Turn-3 prompt:}\\
\texttt{[Turn-2-prompt] + [Turn-2-Response]} \ \\
\texttt{"Please provide a completely different final answer from your previous response, while keeping the same output format."}
\end{quote}
\end{tcolorbox}

\begin{tcolorbox}[
    breakable,
    enhanced,
    left=-0.5cm, right=-0.5cm, top=2pt, bottom=2pt,
    enlarge top by=0.1cm,
    enlarge bottom by=0.1cm,
    title={\hspace{0.5cm} \ours},
    fonttitle=\bfseries\small
]
\begin{quote}\small
\textbf{Prompt:}\\
\texttt{Analyze the question above and generate EXACTLY 3 DISTINCT reasoning paths that lead to DIFFERENT answers or materially different conclusions.}\\
\texttt{Additionally, assign the remaining probability mass to an 'OTHERS' category.}\\
\texttt{Output your response strictly in the following XML format:}

\medskip
\texttt{<response1> Reasoning Path 1 <special-token></response1>}\\
\texttt{<response2> Reasoning Path 2 <special-token></response2>}\\
\texttt{<response3> Reasoning Path 3 <special-token></response3>}\\
\texttt{<response4> OTHERS <special-token></response4>}
\end{quote}
\end{tcolorbox}

\section{Additional Experimental Results}
\label{app:more}

\subsection{Additional Results of \ours on Sequence Classification}
\label{app:additional-results-sc}

\subsubsection{Impact of Distillation Losses}
\label{subsubsec:ablation_distillation_loss}
Our primary methodology utilizes the standard KL-divergence loss as a natural extension of the cross-entropy for knowledge distillation~\cite{hinton2015distilling,shi2022efficacy,xu2024survey}. 
To assess the sensitivity of our approach to these choices, we experimented with alternative distillation loss functions: Reverse KL-divergence~(\textbf{RKL})~\cite{minka2005divergence} and Total Variation Distance~(\textbf{TVD})~\cite{gibbs2002choosing}.
We evaluate these loss functions on the OBQA~\cite{obqa} and WG-S~\cite{wg} datasets, with the results detailed in \Tabref{tab:ablation-losses}.

\begin{table}[h]
    \caption{
        \textbf{Study of Different Distillation Losses for \ours on Sequence Classification.}
    }
    \vspace{-1em}
    \begin{center}
    \resizebox{0.6\linewidth}{!}{%
        \setlength{\tabcolsep}{8pt}
        \begin{tabular}{clccc}
        \toprule[0.12em]
        \textbf{Dataset}
        & \textbf{Loss}
        & \textbf{ACC}~($\uparrow$) 
        & \textbf{ECE}~($\downarrow$) 
        & \textbf{NLL}~($\downarrow$) \\

        \midrule
    
        \multirow{4}{*}[-0.25em]{OBQA}

        & \blob~\cite{wang2024blob}
        & 87.57\scriptsize{$\pm$0.21}
        & 2.91\scriptsize{$\pm$0.92}
        & 0.34\scriptsize{$\pm$0.01}
        \\

        \cmidrule{2-5}
        
        & \ours-KL~(Ours)
        & 88.17\scriptsize{$\pm$0.67} 
        & \textbf{2.55\scriptsize{$\pm$0.12}} 
        & \textbf{0.33\scriptsize{$\pm$0.02}} 
        \\

        & \ours-RKL
        & 87.63\scriptsize{$\pm$1.24} 
        & 2.70\scriptsize{$\pm$0.73} 
        & 0.34\scriptsize{$\pm$0.02} 
        \\

        & \ours-TVD
        & \textbf{88.71\scriptsize{$\pm$0.59}} 
        & 3.21\scriptsize{$\pm$0.44} 
        & 0.34\scriptsize{$\pm$0.01} 
        \\

        \midrule

        \multirow{4}{*}[-0.25em]{WG-S}

        & \blob~\cite{wang2024blob}
        & 76.45\scriptsize{$\pm$0.37}
        & 9.93\scriptsize{$\pm$0.22}
        & 0.58\scriptsize{$\pm$0.00}
        \\

        \cmidrule{2-5}
        
        & \ours-KL~(Ours)
        & \textbf{77.58\scriptsize{$\pm$1.39}} 
        & \textbf{9.24\scriptsize{$\pm$1.16}} 
        & \textbf{0.55\scriptsize{$\pm$0.03}} 
        \\ 

        & \ours-RKL
        & 75.16\scriptsize{$\pm$1.43} 
        & 13.23\scriptsize{$\pm$0.60}
        & 0.63\scriptsize{$\pm$0.02} 
        \\

        & \ours-TVD
        & 51.42\scriptsize{$\pm$0.91}
        & 28.52\scriptsize{$\pm$0.57} 
        & 1.07\scriptsize{$\pm$0.02} 
        \\

        \bottomrule[0.12em]
        \end{tabular}
    }
    \end{center}
    \vspace{-1em}
    \label{tab:ablation-losses}
\end{table}

On OBQA, our standard KL-divergence yields the best ECE (2.55) and NLL (0.33), while TV-Distance achieves slightly higher accuracy (88.71 vs. 88.17 for KL). Reverse-KL shows comparable performance to KL (Ours) on ECE and NLL, with slightly lower accuracy.
On the WG-S dataset, our KL-divergence provides the best ECE (13.00) and NLL (0.62), as well as accuracy. TV-Distance has some interesting results, as the model quickly diverges and gets to an accuracy of 51.2, with might higher ECE and NLL. During this time, the training accuracy, ECE, and NLL continued to drop in the model. TV-Distance is prone to these kinds of divergences and so is not an effective candidate for use as a distillation loss.

Overall, while there are minor variations in performance depending on the specific dataset and metric, no single alternative loss function consistently or significantly outperforms the standard KL-divergence across multiple datasets. The differences are generally small, suggesting that our distillation framework is robust to the precise formulation of the distribution matching loss, as long as it effectively encourages the student to mimic the teacher's predictive distribution. The simplicity and widespread use of KL-divergence make it a suitable default choice for our method.

\begin{figure}[t]
  \centering
  \begin{subfigure}[b]{0.323\textwidth}
    \includegraphics[width=\textwidth]{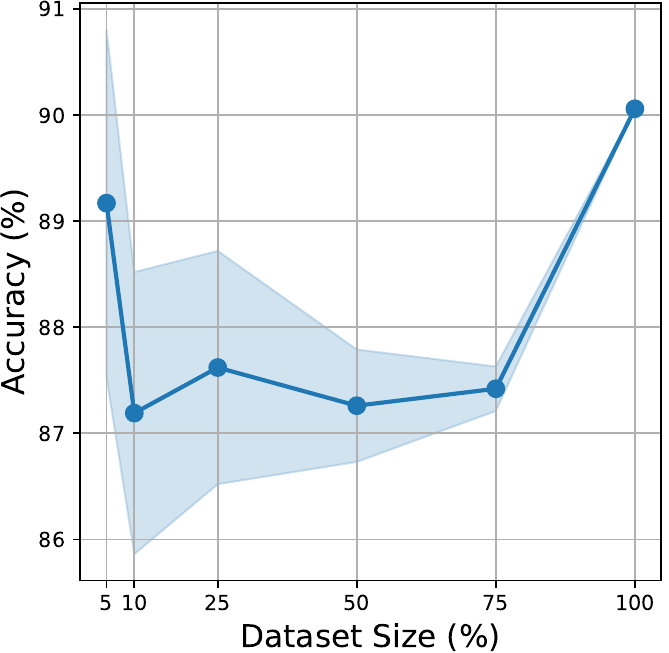}
    \caption{Accuracy}
    \label{fig:boolq_acc}
  \end{subfigure}
  \hfill
  \begin{subfigure}[b]{0.32\textwidth}
    \includegraphics[width=\textwidth]{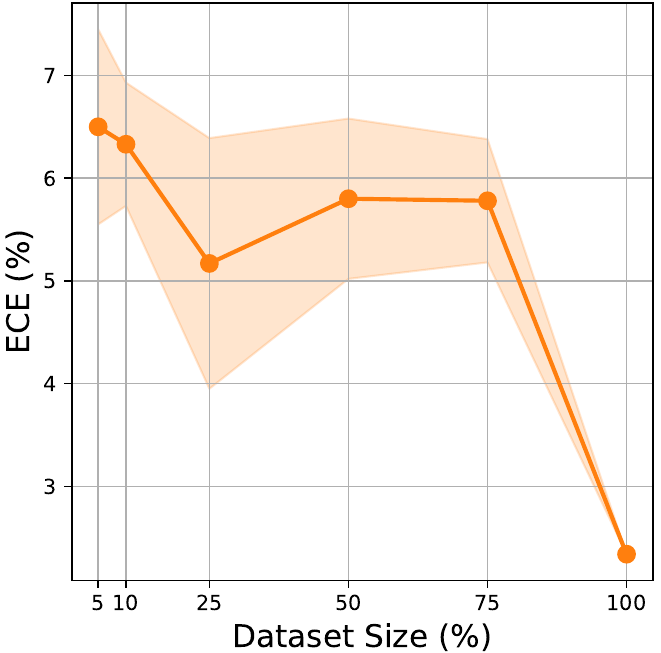}
    \caption{ECE}
    \label{fig:boolq_ece}
  \end{subfigure}
  \hfill
  \begin{subfigure}[b]{0.335\textwidth}
    \includegraphics[width=\textwidth]{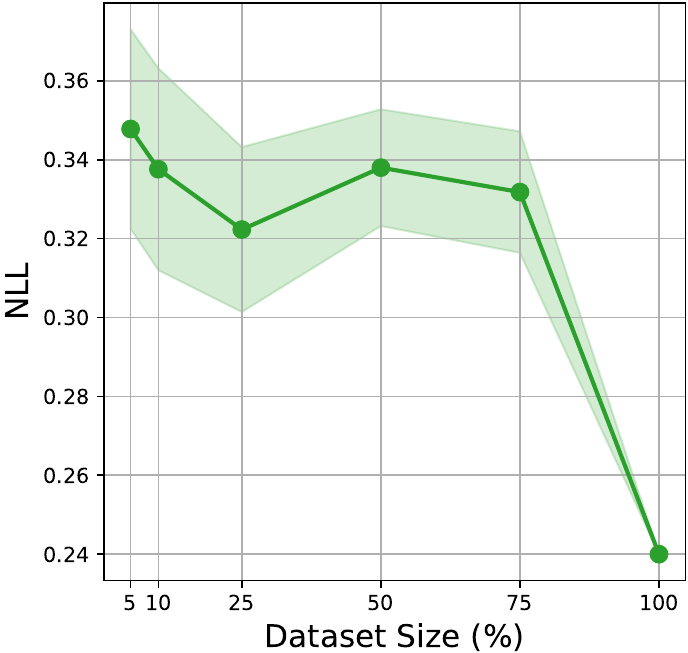}
    \caption{NLL}
    \label{fig:boolq_nll}
  \end{subfigure}
  \caption{
  \textbf{Impact of Dataset Size on Student Model Performance on BoolQ~\cite{boolq}.} 
  Reducing the size of training data leads to higher ECE and NLL, and generally lower ACC, highlighting the importance of sufficient data for effective uncertainty distillation for \ours.
  }
  \label{fig:boolq-metrics}
\end{figure}

\begin{table*}[h]
\caption{
    \textbf{Performance of different LLMs' internal distributions of the answers,} i.e., $\pi_\vtheta(\vy \mid \vx) \triangleq \E_{\vz\sim \pi_\vtheta(\vz\mid\vx)}[\pi_\vtheta(\vy \mid \vz, \vx)]$. Accuracy~(\textbf{ACC}) and Expected Calibration Error~(\textbf{ECE}) are reported in percentages. 
}
\vspace{-1em}
\begin{center}
\resizebox{1\linewidth}{!}{%
\setlength{\tabcolsep}{4pt}
\begin{tabular}{clcccc ccc cccc}
	\toprule[0.12em]
	\multirow{2}{*}[-0.25em]{\textbf{Metric}} 
    & \multirow{2}{*}[-0.25em]{\textbf{LLM Backbone}} 
    & \multicolumn{7}{c}{\textbf{\#Samples for Reasoning Traces $\vz$}}
    \\

    \cmidrule{3-9}
    
    & 
    & 1
    & 3
    & 5
    & 10
    & 20
    & 50
    & 100
    \\
    \midrule
    
    \multirow{8}{*}{ACC~($\uparrow$)} 
    & \gemmaoneb
    & 18.05\scriptsize{$\pm$0.21}
    & 19.32\scriptsize{$\pm$0.21}
    & 20.53\scriptsize{$\pm$0.18}
    & 21.46\scriptsize{$\pm$0.17}
    & 21.96\scriptsize{$\pm$0.14}
    & 22.22\scriptsize{$\pm$0.09}
    & 22.21\scriptsize{$\pm$0.00}
    \\
    
    & \llamaoneb
    & 14.71\scriptsize{$\pm$0.24}
    & 15.79\scriptsize{$\pm$0.21}
    & 17.26\scriptsize{$\pm$0.21}
    & 19.03\scriptsize{$\pm$0.18}
    & 20.08\scriptsize{$\pm$0.15}
    & 20.73\scriptsize{$\pm$0.12}
    & 20.92\scriptsize{$\pm$0.00}
    \\

    & \llamaeightb
    & 23.78\scriptsize{$\pm$0.30}
    & 25.47\scriptsize{$\pm$0.26}
    & 27.59\scriptsize{$\pm$0.23}
    & 29.87\scriptsize{$\pm$0.21}
    & 31.12\scriptsize{$\pm$0.18}
    & 32.00\scriptsize{$\pm$0.13}
    & 32.19\scriptsize{$\pm$0.00}
    \\

    & \qwenthreeb
    & 31.54\scriptsize{$\pm$0.30}
    & 34.15\scriptsize{$\pm$0.29}
    & 36.95\scriptsize{$\pm$0.26}
    & 39.84\scriptsize{$\pm$0.22}
    & 41.51\scriptsize{$\pm$0.19}
    & 42.63\scriptsize{$\pm$0.13}
    & 43.07\scriptsize{$\pm$0.00}
    \\

    & \qwenmathoneb
    & 47.16\scriptsize{$\pm$0.29}
    & 49.82\scriptsize{$\pm$0.27}
    & 52.17\scriptsize{$\pm$0.22}
    & 54.07\scriptsize{$\pm$0.18}
    & 55.06\scriptsize{$\pm$0.21}
    & 55.71\scriptsize{$\pm$0.13}
    & 55.98\scriptsize{$\pm$0.00}
    \\

    & \qwenmathsevenb
    & 48.29\scriptsize{$\pm$0.30}
    & 50.99\scriptsize{$\pm$0.28}
    & 53.52\scriptsize{$\pm$0.23}
    & 55.64\scriptsize{$\pm$0.27}
    & 56.67\scriptsize{$\pm$0.16}
    & 57.25\scriptsize{$\pm$0.12}
    & 57.38\scriptsize{$\pm$0.00}
    \\

    & \qwenfourb
    & 54.54\scriptsize{$\pm$0.30}
    & 58.11\scriptsize{$\pm$0.31}
    & 61.31\scriptsize{$\pm$0.26}
    & 63.82\scriptsize{$\pm$0.22}
    & 65.10\scriptsize{$\pm$0.17}
    & 65.93\scriptsize{$\pm$0.12}
    & 66.29\scriptsize{$\pm$0.00}
    \\

    & \qweneightb
    & 55.53\scriptsize{$\pm$0.30}
    & 59.26\scriptsize{$\pm$0.29}
    & 62.54\scriptsize{$\pm$0.24}
    & 64.99\scriptsize{$\pm$0.20}
    & 66.23\scriptsize{$\pm$0.17}
    & 66.99\scriptsize{$\pm$0.13}
    & 67.23\scriptsize{$\pm$0.00}
    \\

    \midrule
    
    \multirow{8}{*}{ECE~($\downarrow$)} 
    & \gemmaoneb
    & 81.95\scriptsize{$\pm$0.21}
    & 39.94\scriptsize{$\pm$0.25}
    & 33.86\scriptsize{$\pm$0.21}
    & 29.28\scriptsize{$\pm$0.20}
    & 26.85\scriptsize{$\pm$0.15}
    & 25.43\scriptsize{$\pm$0.10}
    & 25.06\scriptsize{$\pm$0.00}
    \\
    
    & \llamaoneb
    & 85.29\scriptsize{$\pm$0.24}
    & 33.67\scriptsize{$\pm$0.23}
    & 24.78\scriptsize{$\pm$0.22}
    & 18.21\scriptsize{$\pm$0.20}
    & 14.70\scriptsize{$\pm$0.17}
    & 12.46\scriptsize{$\pm$0.12}
    & 11.71\scriptsize{$\pm$0.00}
    \\

    & \llamaeightb
    & 76.22\scriptsize{$\pm$0.30}
    & 28.61\scriptsize{$\pm$0.29}
    & 20.22\scriptsize{$\pm$0.25}
    & 13.78\scriptsize{$\pm$0.21}
    & 10.44\scriptsize{$\pm$0.20}
    &  8.26\scriptsize{$\pm$0.13}
    &  7.62\scriptsize{$\pm$0.00}
    \\

    & \qwenthreeb
    & 68.46\scriptsize{$\pm$0.30}
    & 24.78\scriptsize{$\pm$0.29}
    & 16.63\scriptsize{$\pm$0.27}
    &  9.93\scriptsize{$\pm$0.25}
    &  6.29\scriptsize{$\pm$0.20}
    &  4.01\scriptsize{$\pm$0.14}
    &  3.58\scriptsize{$\pm$0.00}
    \\

    & \qwenmathoneb
    & 52.84\scriptsize{$\pm$0.29}
    & 23.47\scriptsize{$\pm$0.27}
    & 18.02\scriptsize{$\pm$0.24}
    & 13.78\scriptsize{$\pm$0.21}
    & 11.59\scriptsize{$\pm$0.21}
    & 10.34\scriptsize{$\pm$0.12}
    &  9.94\scriptsize{$\pm$0.00}
    \\

    & \qwenmathsevenb
    & 51.71\scriptsize{$\pm$0.30}
    & 21.91\scriptsize{$\pm$0.30}
    & 16.20\scriptsize{$\pm$0.25}
    & 11.73\scriptsize{$\pm$0.25}
    &  9.52\scriptsize{$\pm$0.16}
    &  8.25\scriptsize{$\pm$0.13}
    &  7.87\scriptsize{$\pm$0.00}
    \\

    & \qwenfourb
    & 45.46\scriptsize{$\pm$0.30}
    & 16.78\scriptsize{$\pm$0.31}
    & 10.75\scriptsize{$\pm$0.26}
    &  6.25\scriptsize{$\pm$0.21}
    &  5.04\scriptsize{$\pm$0.26}
    &  5.24\scriptsize{$\pm$0.19}
    &  5.37\scriptsize{$\pm$0.00}
    \\

    & \qweneightb
    & 44.47\scriptsize{$\pm$0.30}
    & 16.18\scriptsize{$\pm$0.31}
    & 10.17\scriptsize{$\pm$0.26}
    &  5.93\scriptsize{$\pm$0.22}
    &  5.33\scriptsize{$\pm$0.23}
    &  5.65\scriptsize{$\pm$0.19}
    &  5.71\scriptsize{$\pm$0.00}
    \\

    \midrule
    
    \multirow{8}{*}{NLL~($\downarrow$)} 
    & \gemmaoneb
    & 15.10\scriptsize{$\pm$0.04}
    & 13.67\scriptsize{$\pm$0.03}
    & 13.01\scriptsize{$\pm$0.04}
    & 12.11\scriptsize{$\pm$0.04}
    & 11.22\scriptsize{$\pm$0.04}
    & 10.09\scriptsize{$\pm$0.03}
    &  9.28\scriptsize{$\pm$0.00}
    \\
    
    & \llamaoneb
    & 15.71\scriptsize{$\pm$0.04}
    & 14.31\scriptsize{$\pm$0.04}
    & 13.58\scriptsize{$\pm$0.04}
    & 12.51\scriptsize{$\pm$0.05}
    & 11.41\scriptsize{$\pm$0.04}
    & 10.03\scriptsize{$\pm$0.03}
    &  9.09\scriptsize{$\pm$0.00}
    \\

    & \llamaeightb
    & 14.04\scriptsize{$\pm$0.05}
    & 12.21\scriptsize{$\pm$0.05}
    & 11.31\scriptsize{$\pm$0.04}
    & 10.07\scriptsize{$\pm$0.04}
    &  8.88\scriptsize{$\pm$0.04}
    &  7.48\scriptsize{$\pm$0.03}
    &  6.58\scriptsize{$\pm$0.00}
    \\

    & \qwenthreeb
    & 12.61\scriptsize{$\pm$0.06}
    & 10.18\scriptsize{$\pm$0.05}
    &  9.11\scriptsize{$\pm$0.05}
    &  7.75\scriptsize{$\pm$0.05}
    &  6.59\scriptsize{$\pm$0.03}
    &  5.38\scriptsize{$\pm$0.03}
    &  4.69\scriptsize{$\pm$0.00}
    \\

    & \qwenmathoneb
    &  9.73\scriptsize{$\pm$0.05}
    &  7.45\scriptsize{$\pm$0.04}
    &  6.58\scriptsize{$\pm$0.04}
    &  5.61\scriptsize{$\pm$0.03}
    &  4.84\scriptsize{$\pm$0.03}
    &  4.09\scriptsize{$\pm$0.02}
    &  3.68\scriptsize{$\pm$0.00}
    \\

    & \qwenmathsevenb
    &  9.53\scriptsize{$\pm$0.06}
    &  7.22\scriptsize{$\pm$0.05}
    &  6.35\scriptsize{$\pm$0.04}
    &  5.37\scriptsize{$\pm$0.04}
    &  4.59\scriptsize{$\pm$0.03}
    &  3.84\scriptsize{$\pm$0.02}
    &  3.43\scriptsize{$\pm$0.00}
    \\

    & \qwenfourb
    &  8.37\scriptsize{$\pm$0.06}
    &  5.82\scriptsize{$\pm$0.05}
    &  4.97\scriptsize{$\pm$0.04}
    &  4.13\scriptsize{$\pm$0.03}
    &  3.52\scriptsize{$\pm$0.03}
    &  2.96\scriptsize{$\pm$0.02}
    &  2.66\scriptsize{$\pm$0.00}
    \\

    & \qweneightb
    &  8.19\scriptsize{$\pm$0.05}
    &  5.63\scriptsize{$\pm$0.04}
    &  4.81\scriptsize{$\pm$0.03}
    &  3.99\scriptsize{$\pm$0.03}
    &  3.40\scriptsize{$\pm$0.03}
    &  2.85\scriptsize{$\pm$0.02}
    &  2.56\scriptsize{$\pm$0.00}
    \\
     
    \bottomrule[0.12em]
    \end{tabular}
 }
\end{center}
\label{tab:main-iau}
\vspace{-0.5em}
\end{table*}

\subsubsection{Simple Loss Scheduling.}
\label{app:simple_loss_s}
Directly training the student model with soft targets from the Bayesian teacher can lead to instability. This occurs because finding a single point estimate that matches the representational capacity of an entire approximate distribution $q(\cdot|\vphi)$ is challenging. To stabilize training, we introduce a loss scheduling scheme that initially focuses on learning the mode prediction (highest probability output) before gradually transitioning to full distribution matching. This progressive approach ensures more reliable convergence while maintaining fidelity to the teacher's uncertainty characteristics.
Denote the cross-entropy loss over the true label $\vy^*$ as
\begin{align}
    \gL_{\text{CE}}(\vtheta) 
    &= \E_{(\vx,\vy^*)\sim\gD}\left[ 
        \operatorname{KL}[\delta(\vy\mid\vy^*) | P(\vy\mid\vx, \vtheta)]
    \right], 
\end{align}
where $\delta(\vy|\vy^*)$ is the dirac-delta function assigning full probability to the ground-truth answer $\vy^*$.

At training step $t$, the student model's training loss $\gL_t$ is formulated as a weighted combination of the distillation loss and the cross-entropy loss: 
\begin{align}\label{eq:total}
    \gL_{t}(\vtheta; \vphi) &= \alpha_t \gL_{\text{KD}}(\vtheta; \vphi) + (1-\alpha_t) \gL_{\text{CE}}(\vtheta),
\end{align}
where $\alpha_t=\min(\frac{t}{T_{\alpha}}, 1)$ and $T_{\alpha}$=1,000 is the maximum number of iterations for loss scheduling across all our experiments. 
The rationale for setting $T_{\alpha}$ lower than the total training steps is to ensure the student model's final convergence to the Bayesian teacher's characteristics. With excessively long training at fixed weights, the point estimation (student) would risk overfitting and becoming overconfident about its responses. For a detailed description of \ours, please refer to \Algref{alg:eud}.

\subsubsection{Dataset Augmentation.}
\label{app:data_aug}
Dataset size significantly impacts distillation effectiveness. As demonstrated in \Cref{fig:boolq-metrics}, larger uncertainty distillation datasets substantially improve ECE and NLL metrics, enhancing \ours's uncertainty estimation capabilities.
To address this, we implement dataset augmentation for smaller datasets (WG-S, ARC-C, ARC-E, and WG-M), effectively doubling their size with artificially generated data. We create these additional examples by prompting GPT-4o-mini \cite{gpt4omini} with:

    \begin{tcolorbox}
    {\small
    \emph{``Write a question with the exact same meaning as the one attached, 
    just with different words.''}
    }
    \end{tcolorbox}

We then integrate these paraphrased questions into our complete training pipeline, first having the Bayesian teacher model generate inference outputs on this expanded dataset, then training the student model on this larger collection of teacher outputs.

Importantly, for these synthetic examples, we exclusively use the teacher model's responses for training \ours, \emph{avoiding any unfair advantage} that might come from exposing our method to additional correctly labeled data compared to the teacher model and other baselines.

\subsubsection{Performance of Off-the-Shelf LLMs' Internal Distributions}
\label{app:internal_dist}
The detailed numerical results of the study on \textit{Internal Alignment of Uncertainty~(IAU)} are reported in \Tabref{tab:main-iau}.

\subsubsection{Additional Downstream Adaptation Results on Llama3.1-8B with LoRA}
\label{app:llama_lora_more}
\begin{table*}[h]
\label{llama results}
\caption{
    \textbf{Performance of different methods applied to LoRA on Llama3.1-8B pre-trained weights,} where Accuracy~(\textbf{ACC}) and Expected Calibration Error~(\textbf{ECE}) are reported in percentages. 
    \textbf{``SF?''} stands for whether a method is sampling-free during inference, and we use $N=10$ samples in all sampling-based baseline methods.
    \ours uses \blob~\cite{wang2024blob} as the teacher model and is trained for 10,000 iterations. 
    ``$\uparrow$'' and ``$\downarrow$'' indicate that higher and lower values are preferred, respectively. 
    \textbf{Boldface} and \underline{underlining} denote the best and the second-best performance, respectively.
}
\vspace{-1em}
\begin{center}
\resizebox{1\linewidth}{!}{%
\setlength{\tabcolsep}{8pt}
\begin{tabular}{clcccc ccc}
	\toprule[0.12em]
	\multirow{2}{*}[-0.25em]{\textbf{Metric}} 
    & \multirow{2}{*}[-0.25em]{\textbf{Method}} 
    & \multirow{2}{*}[-0.25em]{\textbf{SF?}}
    & \multicolumn{6}{c}{{\textbf{Dataset}}} 
     \\
     
     \cmidrule(lr){4-9}
     & & & WG-S %
     & ARC-C%
     & ARC-E%
     & WG-M%
     & OBQA
     & BoolQ%

     \\
     \midrule

     \multirow{5}{*}{ACC~($\uparrow$)} 

    & \blob~\cite{wang2024blob} 
    & \textcolor{red}\xmark
    & 76.45\scriptsize{$\pm$0.37}
    & 82.32\scriptsize{$\pm$1.15} 
    & 91.14\scriptsize{$\pm$0.54}  
    & 82.01\scriptsize{$\pm$0.56}  
    & 87.57\scriptsize{$\pm$0.21} 
    & 89.65\scriptsize{$\pm$0.15} 
    \\

    \cmidrule{2-9}

    & \blob-M~\cite{wang2024blob} 
    & \textcolor{Emerald}{\cmark}
    & \textbf{77.72\scriptsize{$\pm$0.12}}  
    & {82.60\scriptsize{$\pm$0.60}}  
    & \textbf{91.64\scriptsize{$\pm$0.55}}  
    & \textbf{83.92\scriptsize{$\pm$0.48}}  
    & 88.00\scriptsize{$\pm$0.80}  
    & 89.86\scriptsize{$\pm$0.05}   
    \\

    & \ours-KL~(Ours)
    & \textcolor{Emerald}{\cmark}
    & \underline{77.58\scriptsize{$\pm$1.39}}  
    & 82.75\scriptsize{$\pm$0.16}
    & \underline{91.37\scriptsize{$\pm$0.33}}  
    & 83.25\scriptsize{$\pm$0.47}  
    & \underline{88.17\scriptsize{$\pm$0.67}}  
    & \underline{90.06\scriptsize{$\pm$0.09}}  
    \\

    & \ours-RKL
    & \textcolor{Emerald}\cmark 
    & 75.16\scriptsize{$\pm$1.43}  
    & \underline{83.22\scriptsize{$\pm$1.43}}
    & 90.43\scriptsize{$\pm$0.44}  
    & \underline{83.60\scriptsize{$\pm$0.79}}  
    & 87.63\scriptsize{$\pm$1.24}  
    & \textbf{90.28\scriptsize{$\pm$0.19}}
    \\

    & \ours-TVD
    & \textcolor{Emerald}\cmark
    & 51.42\scriptsize{$\pm$0.91}  
    & \textbf{83.45\scriptsize{$\pm$0.91}}
    & 69.60\scriptsize{$\pm$29.67}  
    & 83.12\scriptsize{$\pm$0.54}  
    & \textbf{88.71\scriptsize{$\pm$0.59}}  
    & {80.69\scriptsize{$\pm$13.12}}
    \\
     
     \midrule
    
     \multirow{5}{*}{ECE~($\downarrow$)} 
     
    & \blob~\cite{wang2024blob} 
    & \textcolor{red}\xmark
    & 9.93\scriptsize{$\pm$0.22} 
    & 5.41\scriptsize{$\pm$1.17}  
    & 2.70\scriptsize{$\pm$0.87}  
    & 4.28\scriptsize{$\pm$0.64}  
    & 2.91\scriptsize{$\pm$0.92}  
    & 2.58\scriptsize{$\pm$0.25} 
    \\

     \cmidrule{2-9}

    & \blob-M~\cite{wang2024blob} 
    & \textcolor{Emerald}\cmark 
    & 15.43\scriptsize{$\pm$0.15}  
    & 12.41\scriptsize{$\pm$1.52}  
    & 4.91\scriptsize{$\pm$0.28}  
    & 9.37\scriptsize{$\pm$1.33}  
    & 6.44\scriptsize{$\pm$0.15}  
    & 6.26\scriptsize{$\pm$0.29}   
    \\

    & \ours-KL~(Ours)
    & \textcolor{Emerald}\cmark
    & \textbf{9.24\scriptsize{$\pm$1.16}}
    & \textbf{4.76\scriptsize{$\pm$0.58}}
    & \underline{4.49\scriptsize{$\pm$0.56}}
    & \textbf{2.37\scriptsize{$\pm$0.69}}
    & \textbf{2.55\scriptsize{$\pm$0.12}}
    & \textbf{2.34\scriptsize{$\pm$0.28}}
    \\

    & \ours-RKL
    & \textcolor{Emerald}\cmark 
    & \underline{13.23\scriptsize{$\pm$0.60}}  
    & \underline{5.10\scriptsize{$\pm$0.60}}
    & \textbf{2.38\scriptsize{$\pm$0.20}}  
    & \underline{3.52\scriptsize{$\pm$0.43}}  
    & \underline{2.70\scriptsize{$\pm$0.73}}  
    & \underline{2.78\scriptsize{$\pm$0.05}}
    \\

    & \ours-TVD
    & \textcolor{Emerald}\cmark
    & 28.52\scriptsize{$\pm$0.57}  
    & 7.43\scriptsize{$\pm$0.57}
    & 26.15\scriptsize{$\pm$32.68}  
    & 4.64\scriptsize{$\pm$0.47}  
    & 3.21\scriptsize{$\pm$0.44}  
    & {14.77\scriptsize{$\pm$16.34}}
    \\

     \midrule
    
    \multirow{5}{*}{NLL~($\downarrow$)}

    & \blob~\cite{wang2024blob} 
    & \textcolor{red}\xmark 
    & 0.58\scriptsize{$\pm$0.00}  
    & 0.51\scriptsize{$\pm$0.03}  
    & 0.23\scriptsize{$\pm$0.01}  
    & 0.43\scriptsize{$\pm$0.01}  
    & 0.34\scriptsize{$\pm$0.01}  
    & 0.26\scriptsize{$\pm$0.01}  
    \\

    \cmidrule{2-9}

    & \blob-M~\cite{wang2024blob} 
    & \textcolor{Emerald}\cmark
    & 0.74\scriptsize{$\pm$0.02}  
    & 0.73\scriptsize{$\pm$0.04}  
    & 0.29\scriptsize{$\pm$0.03}  
    & 0.47\scriptsize{$\pm$0.03}  
    & 0.37\scriptsize{$\pm$0.02}  
    & 0.32\scriptsize{$\pm$0.02}   
    \\

    & \ours-KL~(Ours)
    & \textcolor{Emerald}\cmark
    & \textbf{0.55\scriptsize{$\pm$0.03}} 
    & \textbf{0.52\scriptsize{$\pm$0.03}} 
    & \textbf{0.24\scriptsize{$\pm$0.01}} 
    & \textbf{0.41\scriptsize{$\pm$0.02}} 
    & \textbf{0.33\scriptsize{$\pm$0.02}} 
    & \textbf{0.24\scriptsize{$\pm$0.00}} 
    \\

    & \ours-RKL
    & \textcolor{Emerald}\cmark 
    & \underline{0.63\scriptsize{$\pm$0.02}}  
    & \textbf{0.52\scriptsize{$\pm$0.02}}
    & \underline{0.26\scriptsize{$\pm$0.01}}  
    & \textbf{0.41\scriptsize{$\pm$0.01}}  
    & \underline{0.34\scriptsize{$\pm$0.02}}  
    & \underline{0.25\scriptsize{$\pm$0.00}}
    \\

    & \ours-TVD
    & \textcolor{Emerald}\cmark
    & 1.07\scriptsize{$\pm$0.02}  
    & 0.55\scriptsize{$\pm$0.02}
    & 4.83\scriptsize{$\pm$6.40}  
    & 0.42\scriptsize{$\pm$0.01}  
    & \underline{0.34\scriptsize{$\pm$0.01}}  
    & {2.30\scriptsize{$\pm$2.88}}
    \\
     
    \bottomrule[0.12em]
    \end{tabular}
 }
\end{center}
\label{tab:ablation-llama-distill-losses}
\vspace{0em}
\end{table*}

\subsection{Additional Results for \ours on Sequence Generation}
\label{app_details_sg}
\subsubsection{The Additional Results of our Trace Cleaning Pipeline for Data Construction}\quad
\label{app:details_1}
\begin{tcolorbox}[
    breakable,
    enhanced,
    left=-0.5cm, right=-0.5cm, top=2pt, bottom=2pt,
    enlarge top by=0.1cm, 
    enlarge bottom by=0.1cm, 
    title={\hspace{0.5cm} Original trace}, 
    fonttitle=\bfseries\small
]
\begin{quote}
We are given:

**Elena drives 45 miles in the first hour**. 
If she continues at the same speed (45 mph), she would be **45 minutes late**. 
She **increases her speed by 20 mph**, so her new speed is **65 mph**.
With this increased speed, she arrives **15 minutes early**.

We need to find the **total distance from Elena's home to the convention center**.

--

\#\#\# Step 1: Let $ D $ be the total distance.

Let $ T $ be the time (in hours) it should take to reach on time.

From the problem:
In the first hour, she travels **45 miles**.
So, the remaining distance is $ D - 45 $ miles.
The remaining time would normally be $ T - 1 $ hours.

If she continued at 45 mph, the time to cover the remaining distance would be:
$$
\frac{D - 45}{45} \text{hours}
$$
So the **total time** would be:
$$
1 + \frac{D - 45}{45} = \frac{D - 45}{45} + 1 = \frac{D - 45 + 45}{45} = \frac{D}{45}
$$
But she is **45 minutes late**, which is $ \frac{3}{4} $ hours.
So:
$$
\frac{D}{45} = T + \frac{3}{4}
$$
Now consider the second scenario where she increases her speed after the first hour.
She now goes at **65 mph** for the remaining distance $ D - 45 $.
Time taken for the remaining distance:
$$
\frac{D - 45}{65} \text{ hours}
$$
Total time is:
$$
1 + \frac{D - 45}{65}
$$
\\This time is **15 minutes early**, i.e., $ T - \frac{1}{4} $
So we have another equation:
$$
1 + \frac{D - 45}{65} = T - \frac{1}{4}
$$

--

\#\#\# Step 2: Use both equations to solve for $ D $
\\From earlier:
$$
\frac{D}{45} = T + \frac{3}{4} (1)
$$
$$
1 + \frac{D - 45}{65} = T - \frac{1}{4} (2)
$$
Subtract (2) from (1):
$$
\left( \frac{D}{45} \right) - \left( 1 + \frac{D - 45}{65} \right) = \left( T + \frac{3}{4} \right) - \left( T - \frac{1}{4} \right)
$$
Left side:
$$
\frac{D}{45} - 1 - \frac{D - 45}{65}
$$
Right side:
$$T + \frac{3}{4} - T + \frac{1}{4} = 1
$$

So:
$$
\frac{D}{45} - 1 - \frac{D - 45}{65} = 1
$$

Add 1 to both sides:
$$
\frac{D}{45} - \frac{D - 45}{65} = 2
$$
--
\#\#\# Step 3: Solve the equation

Find common denominator of 45 and 65, which is 585.

Convert fractions:
$$
\frac{D}{45} = \frac{13D}{585}, \quad \frac{D - 45}{65} = \frac{9(D - 45)}{585}
$$

So:
$$
\frac{13D - 9(D - 45)}{585} = 2
$$

Simplify numerator:
$$
13D - 9D + 405 = 4D + 405
$$

So:
$$
\frac{4D + 405}{585} = 2
$$

Multiply both sides by 585:
$$
4D + 405 = 1170
$$

Subtract 405:
$$
4D = 765
$$

Divide:
$$
D = \frac{765}{4} = \boxed{191.25}
$$
--
\#\#\# \u2705 Final Answer:

$$
\boxed{191.25} \text{ miles}
$$

This is the total distance from Elena's home to the convention center.
\end{quote}
\end{tcolorbox}

\begin{tcolorbox}[
    breakable,
    enhanced,
    left=-0.5cm, right=-0.5cm, top=2pt, bottom=2pt,
    enlarge top by=0.1cm, 
    enlarge bottom by=0.1cm, 
    title={\hspace{1cm} Cleaned trace}, 
    fonttitle=\bfseries\small
]
\begin{quote}
We are given:\\ 
Elena covers 45 miles in the first hour. If she continued at 45 mph she would arrive 45 minutes (= 3/4 hr) late. If after the first hour she goes 20 mph faster (65 mph) she arrives 15 minutes (= 1/4 hr) early. Find the total distance D.\\
Step 1: Let D be total distance and T the on-time travel time (hours).\\
After the first hour the remaining distance is D - 45 and the remaining scheduled time is T - 1.\\
Continuing at 45 mph gives total time\\
1 + (D - 45)/45 = D/45,\\
and this equals T + 3/4:\\
D/45 = T + 3/4.  (1)\\
Going at 65 mph after the first hour gives total time\\
1 + (D - 45)/65,\\
and this equals T $-$ 1/4:\\
1 + (D $-$ 45)/65 = T $-$ 1/4.  (2)\\
Step 2: Eliminate T by subtracting (2) from (1):\\
(D/45) $-$ [1 + (D $-$ 45)/65] = (T + 3/4) $-$ (T $-$ 1/4) = 1.\\
So D/45 $-$ 1 $-$ (D $-$ 45)/65 = 1,\\
hence\\
D/45 $-$ (D $-$ 45)/65 = 2.\\
Step 3: Solve for D. With common denominator 585:\\
13D/585 $-$ 9(D $-$ 45)/585 = $\to$ [13D $-$ 9(D $-$ 45)]/585 = $\to$ (13D $-$ 9D + 405)/585 = $\to$ (4D + 405) = 1170$\to$ 4D = 765$\to$ D = 765/4 = 191.25.\\
Final Answer: \\boxed{191.25}"
\end{quote}
\end{tcolorbox}

\begin{tcolorbox}[
    breakable,
    enhanced,
    left=-0.5cm, right=-0.5cm, top=2pt, bottom=2pt,
    enlarge top by=0.1cm, 
    enlarge bottom by=0.1cm, 
    title={\hspace{0.5cm} Original trace 2}, 
    fonttitle=\bfseries\small
]
\begin{quote}
We are given:

A number $ n $ with **at least 7 positive divisors**.
The **divisors** of $ n $ are ordered as:
$$
1 = d_1 < d_2 < d_3 < d_4 < d_5 < d_6 < d_7 < \cdots < d_k = n
$$
And we are told:
$$
d_7 = 2d_5 + 1 \quad \text{(Equation 1)}
$$
$$
d_7 = 3d_4 - 1 \quad \text{(Equation 2)}
$$

Our goal is to find the **smallest such positive integer $ n $**.

---

\#\#\# Step 1: Equating the two expressions for $ d_7 $

From Equation 1 and Equation 2, we have:
$$
2d_5 + 1 = 3d_4 - 1
$$
$$
2d_5 = 3d_4 - 2
$$
$$
d_5 = \frac{3d_4 - 2}{2}
$$

Since $ d_5 $ must be an integer (it's a divisor), then $ 3d_4 - 2 $ must be even. So,
$$
3d_4 \equiv 2 \pmod{2} \Rightarrow d_4 \equiv 0 \pmod{2}
$$
So, $ d_4 $ is **even**.

Let\u2019s denote:
$$
d_4 = x \Rightarrow d_5 = \frac{3x - 2}{2}
$$
Also, from Equation 2:
$$
d_7 = 3x - 1
$$

Now, recall that the divisors are in increasing order:
$$
1 = d_1 < d_2 < d_3 < d_4 = x < d_5 = \frac{3x - 2}{2} < d_6 < d_7 = 3x - 1
$$

Thus, the first six divisors must satisfy this ordering.

---

\#\#\# Step 2: Try small values of $ x $ (i.e., $ d_4 $) and check if the sequence of divisors works.

Try $ x = 2 $:

Then:
- $ d_4 = 2 $
- $ d_5 = \frac{3(2) - 2}{2} = \frac{6 - 2}{2} = 2 $
$/to$ Not valid since $ d_4 = d_5 $. Invalid.

Try $ x = 4 $:
- $ d_4 = 4 $
- $ d_5 = \frac{3(4) - 2}{2} = \frac{12 - 2}{2} = 5 $
- $ d_7 = 3(4) - 1 = 11 $

Check if there exists a number $ n $ whose first 7 divisors are:
$$
1, d_2, d_3, 4, 5, d_6, 11
$$

This means $ n $ must be divisible by 1, 4, 5, and 11.

Find LCM(1, 4, 5, 11) = LCM(4, 5, 11) = 220

Check all divisors of 220:

Divisors of 220:
$$
1, 2, 4, 5, 10, 11, 20, 22, 44, 55, 110, 220
$$
Order them:
$$
1, 2, 4, 5, 10, 11, 20, 22, 44, 55, 110, 220
$$

So:
- $ d_1 = 1 $
- $ d_2 = 2 $
- $ d_3 = 4 $
- $ d_4 = 5 $
- $ d_5 = 10 $
- $ d_6 = 11 $
- $ d_7 = 20 $

But wait \u2014 we expected $ d_4 = 4 $, but here it's 5.

So this does not match our earlier assumption that $ d_4 = x = 4 $, so invalid.

Try next value $ x = 6 $:

- $ d_4 = 6 $
- $ d_5 = \frac{3(6) - 2}{2} = \frac{18 - 2}{2} = 8 $
- $ d_7 = 3(6) - 1 = 17 $

Now try to find a number $ n $ divisible by 1, 6, 8, 17.

LCM(1, 6, 8, 17) = LCM(6, 8, 17) = 408

Divisors of 408:
$$
1, 2, 3, 4, 6, 8, 12, 17, 24, 34, 51, 68, 102, 136, 204, 408
$$

Order them:
$$
1, 2, 3, 4, 6, 8, 12, 17, ...
$$

So:
- $ d_1 = 1 $
- $ d_2 = 2 $
- $ d_3 = 3 $
- $ d_4 = 4 $
- $ d_5 = 6 $
- $ d_6 = 8 $
- $ d_7 = 12 $

But we needed:
- $ d_4 = 6 $
- $ d_5 = 8 $
- $ d_7 = 17 $

Doesn't fit.

Try $ x = 8 $:

- $ d_4 = 8 $
- $ d_5 = \frac{3(8) - 2}{2} = \frac{24 - 2}{2} = 11 $
- $ d_7 = 3(8) - 1 = 23 $

Try LCM(1, 8, 11, 23) = 2024

Divisors of 2024:

Prime factorization:
$$
2024 = 2^3 \cdot 11 \cdot 23
$$

List of divisors:
$$
1, 2, 4, 8, 11, 22, 23, 44, 46, 88, 92, 184, 253, 506, 1012, 2024
$$

Order them:
$$
1, 2, 4, 8, 11, 22, 23, 44, 46, 88, 92, 184, 253, 506, 1012, 2024
$$

So:
- $ d_1 = 1 $
- $ d_2 = 2 $
- $ d_3 = 4 $
- $ d_4 = 8 $
- $ d_5 = 11 $
- $ d_6 = 22 $
- $ d_7 = 23 $

\u2705 This matches our required structure!

Check the conditions:

- $ d_7 = 23 $
- $ d_5 = 11 $
- $ d_4 = 8 $

Check equations:

- $ d_7 = 2d_5 + 1 = 2(11) + 1 = 23 $ \u2705
- $ d_7 = 3d_4 - 1 = 3(8) - 1 = 23 $ \u2705

All conditions satisfied.

---

\#\#\# Final Answer:

The **smallest positive integer $ n $** satisfying the conditions is:

$$
\boxed{2024}
$$"
\end{quote}
\end{tcolorbox}

\begin{tcolorbox}[
    breakable,
    enhanced,
    left=-0.5cm, right=-0.5cm, top=2pt, bottom=2pt,
    enlarge top by=0.1cm, 
    enlarge bottom by=0.1cm, 
    title={\hspace{1cm} Cleaned trace 2}, 
    fonttitle=\bfseries\small
]
\begin{quote}
We are given n with at least 7 positive divisors, ordered
1 = d1 < d2 < d3 < d4 < d5 < d6 < d7 < ... < dk = n,
and
d7 = 2d5 + 1,    (1)
d7 = 3d4 - 1.    (2)

Step 1: Equate the two expressions for d7.

From (1) and (2):
2d5 + 1 = 3d4 - 1 
21d2 2d5 = 3d4 - 2 
21d2 d5 = (3d4 - 2)/2.

Thus 3d4 - 2 is even, so d4 is even. Let d4 = x, then
d5 = (3x - 2)/2,   d7 = 3x - 1,
with ordering 1 < d2 < d3 < x = d4 < (3x - 2)/2 = d5 < d6 < 3x - 1 = d7.

Step 2: Test small even x for which d5 is integer and divisors can be arranged as required.

x = 2: d5 = (6 - 2)/2 = 2 $/to$ d4 = d5, invalid.

x = 4: d5 = 5, d7 = 11. Any n must be divisible by 1,4,5,11, so n is multiple of lcm(4,5,11)=220. Divisors of 220 ordered: 1,2,4,5,10,11,... so d4 = 4 would be the third or fourth position? Here the sequence gives d4 = 5, not 4. So mismatch.

x = 6: d5 = 8, d7 = 17. n must be multiple of lcm(6,8,17)=408. Divisors of 408 start 1,2,3,4,6,8,12,17,... so d4 = 4 not 6. Mismatch.

x = 8: d5 = 11, d7 = 23. n must be multiple of lcm(8,11,23)=2024. Factorization 2024 = 2$^3$$\cdot$11$\cdot$23. Its divisors in order:
1,2,4,8,11,22,23,44,...
Thus
d1=1, d2=2, d3=4, d4=8, d5=11, d6=22, d7=23,
which matches the required ordering and both equations:
2d5 + 1 = 2$\cdot$11 + 1 = 23, 3d4 - 1 = 3$\cdot$8 - 1 = 23.

Hence the smallest such n is 2024.

Final Answer: \\boxed{2024}"
\end{quote}
\end{tcolorbox}

\begin{tcolorbox}[
    breakable,
    enhanced,
    left=-0.5cm, right=-0.5cm, top=2pt, bottom=2pt,
    enlarge top by=0.1cm, 
    enlarge bottom by=0.1cm, 
    title={\hspace{0.5cm} Original trace 3}, 
    fonttitle=\bfseries\small
]
\begin{quote}
We are given:
- There are **two teams** of **5 runners each**, so **10 runners total**.
- Runners finish in positions from **1st to 10th** (no ties).
- Each runner's contribution to their team's score is equal to their finishing position.
- The **team with the lower total score wins**.
- We are asked: **How many different winning scores are possible?**

---

\#\#\# Step 1: Understand the scoring system

Each runner contributes their position number to their team's score.
Let\u2019s denote:
- Team A has runners in positions $ a_1 < a_2 < a_3 < a_4 < a_5 $
- Team B has runners in positions $ b_1 < b_2 < b_3 < b_4 < b_5 $

Then:
- Team A's score = $ a_1 + a_2 + a_3 + a_4 + a_5 $
- Team B's score = $ b_1 + b_2 + b_3 + b_4 + b_5 $

Since all positions are unique integers from 1 to 10, the sum of all 10 positions is fixed:

$$
\text{Total score} = 1 + 2 + 3 + \dots + 10 = \frac{10 \cdot 11}{2} = 55
$$

So, if Team A has a score $ S_A $, then Team B must have a score $ S_B = 55 - S_A $. Therefore, the **winning score** for the winner is either $ S_A $ or $ S_B $, whichever is smaller.

But since we're only interested in **how many different winning scores are possible**, we need to find all values that can be the **smaller** of two such sums.

That is, we want to count the number of distinct values of $ S_A $ such that $ S_A < 55 - S_A $, i.e., $ S_A < 27.5 $. So any valid $ S_A $ must be less than 28.

Thus, we are looking for the number of distinct values of $ S_A $ where $ S_A $ is the sum of any 5 distinct integers from 1 to 10, and $ S_A < 28 $.

---

\#\#\# Step 2: Find all possible team scores

We now need to compute all possible values of $ S_A $, the sum of any 5 distinct integers from 1 to 10, and count how many of them are **less than 28**.

The minimum possible value of $ S_A $ is when the five smallest numbers are chosen:

$$
1 + 2 + 3 + 4 + 5 = 15
$$

The maximum possible value of $ S_A $ is when the five largest numbers are chosen:

$$
6 + 7 + 8 + 9 + 10 = 40
$$

So possible values of $ S_A $ range from 15 to 40.

Now, we want to count how many of these values are **less than 28**.

That is, we want the number of integer values of $ S_A $ such that:

$$
15 \leq S_A < 28
$$

These values are:

$$
15, 16, 17, 18, 19, 20, 21, 22, 23, 24, 25, 26, 27
$$

That's **13** different values.

Therefore, **there are 13 different winning scores** possible.

---

\#\#\# $\checkmark$ Final Answer:

$$
\boxed{13}
$$

**(B)** is the correct choice.
\end{quote}
\end{tcolorbox}

\begin{tcolorbox}[
    breakable,
    enhanced,
    left=-0.5cm, right=-0.5cm, top=2pt, bottom=2pt,
    enlarge top by=0.1cm, 
    enlarge bottom by=0.1cm, 
    title={\hspace{0.5cm} Cleaned trace 3}, 
    fonttitle=\bfseries\small
]
\begin{quote}
Step 1: Understand the scoring system

Each runner contributes their finishing position to their team's score. Let Team A have positions a1<a2<a3<a4<a5 and Team B have the remaining five positions. Then
Team A's score $S_A = a1+a2+a3+a4+a5$,
Team B's score $S_B = 55 - S_A$,
since $1+2+...+10 = 55$.

The winning score is the smaller of $S_A$ and $S_B$. For Team A to be the winner we need $S_A < S_B, i.e. S_A < 27.5$, so any winning score must satisfy $S < 28$.

Step 2: Find possible team scores

$S_A$ is the sum of any 5 distinct integers from 1 to 10. The minimum is $1+2+3+4+5 = 15$ and the maximum is $6+7+8+9+10 = 40$. Thus possible sums range from 15 to 40. We want those less than 28, i.e. the integers
15,16,17,18,19,20,21,22,23,24,25,26,27,
which are 13 distinct values.

Final Answer: \boxed{13}
\end{quote}
\end{tcolorbox}

\subsubsection{2. A Qualitative Comparison between \ours and Baseline Methods}\quad
\label{app:comparsion}
We further present a qualitative example comparing \ours with two baselines on the same problem. 
In this specific case, while all methods failed to output the correct answer, the calibration metrics derived from the aggregated answer probabilities reveal a significant difference in uncertainty estimation.
The baselines exhibited severe overconfidence in their incorrect predictions, yielding high Top-1 ECE scores of 0.67 and 1.00. 
In contrast, \ours demonstrates superior calibration performance, achieving significantly lower Top-1 and class wise ECE scores of 0.45 and 0.25. 
This result highlights that \ours maintains a more reliable confidence estimate, avoiding the extreme overconfidence observed in the baselines even when the prediction is inaccurate.

\begin{tcolorbox}[
    breakable,
    enhanced,
    left=0.1cm, right=0.1cm, top=2pt, bottom=2pt,
    enlarge top by=0.1cm,
    enlarge bottom by=0.1cm,
    title={\centering Qualitative Comparison},
    fonttitle=\bfseries\small,
    colback=white,
    colframe=gray!75!black
]
\small
\textbf{Problem:} \\
\textit{Question: In the circle with center $Q$, radii $AQ$ and $BQ$ form a right angle. The two smaller regions are tangent semicircles, as shown. The radius of the circle with center $Q$ is 14 inches. What is the radius of the smaller semicircle? Express your answer as a common fraction.}

\begin{description}[leftmargin=1.2cm, style=nextline]
    \item[\textbf{VS (Vanilla Sampling):}] 
    ``<response> \\
    The radius of the smaller semicircle is $\boxed{\frac{7}{2}}$ <probability>0.65probs<\textbackslash probability> \\
    </response> \\
    \\
    <response> \\
    The radius of the smaller semicircle is $\boxed{\frac{7}{2}}$ <probability>0.75probs<\textbackslash probability> \\
    </response> \\
    \\
    <response> \\
    The radius of the smaller semicircle is $\boxed{\frac{7}{2}}$ <probability>0.85probs<\textbackslash probability> \\
    </response>''

    \medskip
    \item[\textbf{MT (Multiturn-Sampling):}]{ 
    \textbf{Turn-1:} We are given a circle with center $Q$ and radius 14 inches. Two radii, $AQ$ and $BQ$, form a right angle, so triangle $AQB$ is a right triangle with right angle at $Q$. Inside this circle, there are **two smaller tangent semicircles**, and we are to find the **radius of one of these smaller semicircles**. Since $AQ$ and $BQ$ are perpendicular, the triangle is a **right isosceles triangle** with legs of length 14, and hypotenuse: $AB = \sqrt{14^2 + 14^2} = \sqrt{196 + 196} = \sqrt{392} = 14\sqrt{2}$. Let's place the circle with center at the origin $(0, 0)$, and let: $A = (14, 0)$, $B = (0, 14)$. Then the semicircles are tangent to both axes (the radii $AQ$ and $BQ$), and also tangent to the arc of the main circle. Let the center of one of the smaller semicircles be at point $(r, r)$, where $r$ is the radius of the semicircle. This semicircle is tangent to both the x-axis and y-axis, so its center must be at $(r, r)$, and its radius is $r$. Now, the semicircle is also **tangent to the main circle**. The main circle has center $(0, 0)$ and radius 14. The distance between the centers is: $\sqrt{(r - 0)^2 + (r - 0)^2} = \sqrt{2r^2} = r\sqrt{2}$. Since the semicircle is **inside** the main circle, the distance between the centers is the **difference** of the radii: $r\sqrt{2} = 14 - r$. Solve: $r\sqrt{2} + r = 14 \Rightarrow r(\sqrt{2} + 1) = 14$. $r = \frac{14}{\sqrt{2} + 1}$. Rationalize: $r = \frac{14(\sqrt{2} - 1)}{2 - 1} = 14(\sqrt{2} - 1)$. Final answer: $\boxed{14(\sqrt{2} - 1)}$. \\
    \textbf{Turn-2:} We are given a circle with center $Q$ and radius 14 inches. Two radii, $AQ$ and $BQ$, form a right angle, so triangle $AQB$ is a right isosceles triangle with legs of length 14, and hypotenuse: $AB = \sqrt{14^2 + 14^2} = 14\sqrt{2}$. We are to find the **radius of one of two smaller tangent semicircles**. Assume the smaller semicircles are tangent to each other and to the two radii $AQ$ and $BQ$, and also tangent to the arc of the main circle. Let the center of the main circle be at $(0, 0)$, with $A = (14, 0)$ and $B = (0, 14)$. Let the center of one smaller semicircle be at point $(r, r)$, where $r$ is the radius. The distance between the centers is: $\sqrt{(r - 0)^2 + (r - 0)^2} = r\sqrt{2}$. Since the semicircle is **inside** the main circle, the distance between the centers is the **difference** of the radii: $r\sqrt{2} = 14 - r$. Solve: $r(\sqrt{2} + 1) = 14 \Rightarrow r = \frac{14}{\sqrt{2} + 1}$. Rationalize: $r = \frac{14(\sqrt{2} - 1)}{1} = 14(\sqrt{2} - 1)$. This is less than 14, so it's valid. Final answer: $\boxed{\frac{14(\sqrt{2} - 1)}{1}}$. \\
    \textbf{Turn-3:} We are given a circle with center $Q$ and radius 14 inches. Two radii, $AQ$ and $BQ$, form a right angle, so triangle $AQB$ is a right isosceles triangle with legs of length 14 and hypotenuse $AB = 14\sqrt{2}$. Inside this circle, there are **two smaller tangent semicircles**. Let the center of the main circle be $(0, 0)$ and the center of one smaller semicircle be $(r, r)$, where $r$ is its radius. The distance between the centers is $r\sqrt{2}$. Because the semicircle is tangent to the main circle internally, we have the relationship $r\sqrt{2} + r = 14$. Solving for $r$: $r(\sqrt{2} + 1) = 14 \Rightarrow r = \frac{14}{\sqrt{2} + 1}$.  Final answer: $\boxed{\frac{14}{\sqrt{2} + 1}}$.}

    \medskip
    \item[\textbf{Ours:}] 
    <response1> 
    We place the circle center at the origin $(0,0)$. The small semicircle has its center at $(r,r)$, so the distance to the origin is $r\sqrt{2}$. Assuming the tangency relationship is $r\sqrt{2} = 14 + r$, we solve for $r$ to get $r(\sqrt{2}-1) = 14$, which implies $r = 14(\sqrt{2}+1)$. Final Answer: $\boxed{14(\sqrt{2}+1)}$ <special-token>
    </response1> \\
    <response2> 
    We use half of the main circle's radius, so $r = \frac{14}{2} = 7$. Final Answer: $\boxed{7}$ <special-token>
    </response2> \\
    <response3> 
    The radius is $\frac{7}{2}$. Final Answer: $\boxed{\frac{7}{2}}$ <special-token>
    </response3> \\
    <response4> 
    $\boxed{OTHERS}$ <special-token>
    </response4>

\end{description}
\end{tcolorbox}

\end{document}